\documentclass[conference]{IEEEtran}
\usepackage{times}

\usepackage[numbers]{natbib}
\usepackage{multicol}
\usepackage[bookmarks=true]{hyperref}

\usepackage{graphicx}
\usepackage{makecell}
\usepackage{amsmath}
\usepackage{amsfonts}
\usepackage{wrapfig}
\usepackage{caption}
\usepackage{booktabs}
\usepackage{subcaption}
\usepackage[table]{xcolor}   

\definecolor{myblue}{rgb}{0.88,0.98,1} 
\definecolor{deemph}{gray}{0.6}

\newcommand{\gc}[1]{\textcolor{deemph}{#1}}
\newcommand{\ourdataset}{Dex1B}
\newcommand{\ourmethod}{DexSimple}

\makeatletter
\def\blfootnote{\xdef\@thefnmark{}\@footnotetext}
\makeatother

\begin{document}

\title{\ourdataset: Learning with 1B Demonstrations for Dexterous Manipulation}

\author{\authorblockN{Jianglong Ye$^*$, Keyi Wang$^*$, Chengjing Yuan, Ruihan Yang, Yiquan Li, Jiyue Zhu\\Yuzhe Qin, Xueyan Zou, Xiaolong Wang}
\vspace{0.05in}
\authorblockA{UC San Diego}
{\texttt{\url{https://jianglongye.com/dex1b}}}
}

\twocolumn[{%
\renewcommand\twocolumn[1][]{#1}%
\maketitle
\begin{center}
    \vspace{-0.15in}
    \centering
    \captionsetup{type=figure}
    \includegraphics[width=1.0\linewidth]{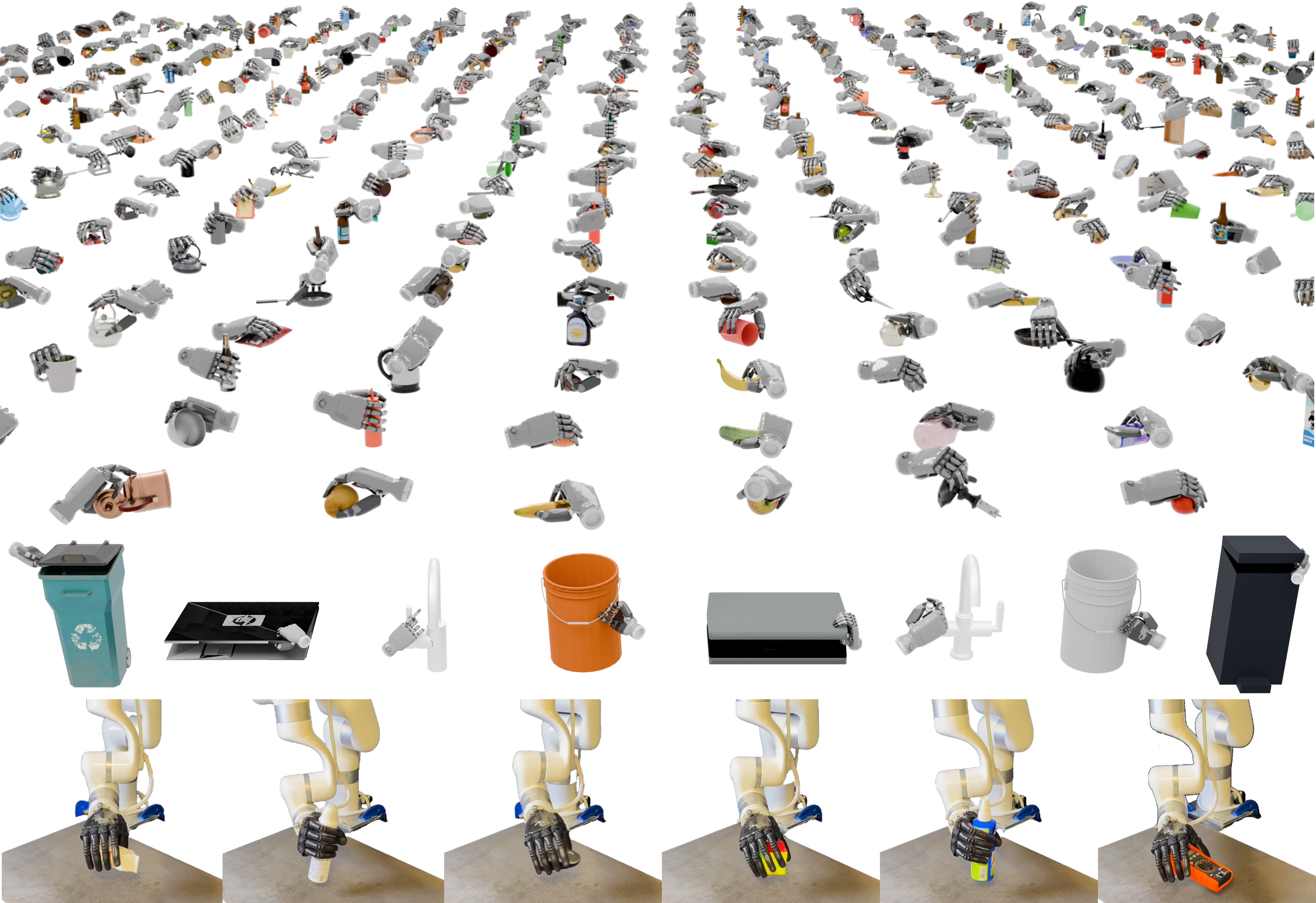}
    \vspace{-0.2in}
    \captionof{figure}{The \textbf{\ourdataset} benchmark consists of 1B generated high-quality demonstrations for grasping (top) and articulation (middle) tasks. At the bottom, we show the \textbf{direct sim-to-real} transfer results of our method \textbf{\ourmethod}~trained on \ourdataset. This demonstrates that \ourdataset~is scalable and generalizable to real environments.}
    \label{fig:teaser}
\end{center}
}]
{\blfootnote{{$^{*}$ Equal contribution.}}}


\begin{abstract}
Generating large-scale demonstrations for dexterous hand manipulation remains challenging, and several approaches have been proposed in recent years to address this. Among them, generative models have emerged as a promising paradigm, enabling the efficient creation of diverse and physically plausible demonstrations.
In this paper, we introduce \ourdataset, a large-scale, diverse, and high-quality demonstration dataset produced with generative models. The dataset contains  one billion demonstrations for two fundamental tasks: grasping and articulation.
To construct it, we propose a generative model that integrates geometric constraints to improve feasibility and applies additional conditions to enhance diversity.
We validate the model on both established and newly introduced simulation benchmarks, where it significantly outperforms prior state-of-the-art methods. Furthermore, we demonstrate its effectiveness and robustness through real-world robot experiments.
\end{abstract}

\IEEEpeerreviewmaketitle

\vspace{-5pt}
\section{Introduction}
\vspace{-5pt}

Dexterous manipulation with hand has been a long-standing topic in robotics. While its highly flexible and dynamic nature allows for more complex and robust manipulation skills, the high degrees of freedom (DoF) of a hand makes it very challenging to achieve its ideal function. In fact, with recent advancements in applications using parallel-jaw grippers~\cite{DBLP:conf/rss/ZhaoKLF23,fu2024mobile,chi2024universal,zhao2024aloha,black2024pi_0}, researchers in the community have started questioning the necessity of dexterous hands and having doubts about whether hands are only making problems harder.

We argue that dexterous hand is indeed valuable, but we just did not have enough data to capture the diverse and complex distributions required for effective dexterous manipulation. To address this data scarcity, previous works have explored various approaches, including human demonstrations~\cite{DBLP:conf/ijcai/LiuYWWWYSYWY0M24,qin2022dexmv,chen2022dextransfer,chen2024object}, optimization-based methods~\cite{wang2023dexgraspnet,chen2023task,turpin2023fast}, reinforcement learning (RL)-based techniques~\cite{christen2024synh2r,zhao2024rp1m}. While these methods help generate demonstrations at a certain scale, they each have limitations: human annotation is costly and imprecise, optimization-based methods are slow and sensitive to initialization, and RL-based techniques lack data diversity.

Meanwhile, a large body of generative models~\cite{jiang2021graspTTA,lu2023ugg,ye2023learning,liu2023contactgen,weng2024DexDiffuser} has been proposed in recent years to model the distribution of demonstration datasets, motivating us to explore how generative models can be leveraged for data generation. And we identify two key issues when applying generative models on data generation:  i). \textbf{Feasibility:} The success rate of generative models is often lower than that of deterministic models. ii). \textbf{Diversity:} While generative models can produce more diverse actions than deterministic  models, they still tend to interpolate between the seen demonstrations, which may maintain or even reduce the original level of diversity of whole dataset rather than expanding it.

To address the feasibility issue, we propose incorporating geometric constraints into the generative model, which significantly improves its performance. We also integrate optimization techniques with generative models, leveraging the strengths of both approaches: optimization ensures physically plausible results, while generative models enable efficient large-scale generation. To improve diversity, we introduce additional conditions to the generative model and prioritize sampling actions for less frequent condition values, encouraging the model to generate actions that differ more from existing ones in the dataset.

Our approach begins with an optimization-based method to construct a small yet high-quality seed dataset of dexterous manipulation demonstrations. We then train a generative model on this dataset and use it to scale up data generation efficiently. To mitigate biases introduced by optimization, we propose an debias mechanism, which systematically improves the diversity of generated data. This framework results in \ourdataset, a dataset comprising one billion dexterous hand demonstrations, representing a substantial advancement in scale, diversity and quality over existing datasets.
Compared to DexGraspNet~\cite{wang2023dexgraspnet}, which operates on the object set of similar scale, our dataset offers \textit{$700\times$} more demonstrations, significantly enriching the available training data for learning-based models. Unlike previous approaches that rely solely on human annotation or optimization, our method combines optimization and neural networks, achieving a superior balance between cost, efficiency, and data quality.

To effectively leverage the scale and diversity of \ourdataset, we introduce \ourmethod, a new baseline that extends prior work~\cite{jiang2021graspTTA} by incorporating conditional generation and enhanced loss functions. Despite its simplicity, \ourmethod\ benefits from the scale and diversity of \ourdataset, achieving state-of-the-art (SoTA) performance across dexterous manipulation tasks.

Our key contributions are as follows:
\begin{itemize}
    \item We introduce a novel iterative data generation pipeline that combines optimization and generative models to generate large-scale dexterous demonstrations for grasping and articulation tasks. Using this approach, we construct \ourdataset, the largest and most diverse dexterous demonstration dataset to date.
    \item We propose a simple yet effective baseline method that incorporates enhanced loss functions while supporting conditional generation, making it particularly well-suited for our iterative pipeline and policy deployment.
    \item Leveraging \ourdataset\ and \ourmethod, we achieve a $22\%$ performance improvement over previous best methods on grasp synthesis tasks, setting a new benchmark in dexterous manipulation.
\end{itemize}

\begin{table}[t]
\small
\centering
\setlength{\tabcolsep}{2pt}
\fontsize{8pt}{9pt}\selectfont
\begin{tabular}{l cccc }
\toprule
Dataset & Task  & \makecell{Num \\ Objects} & \makecell{Num Demo-\\nstrations} & Method \\ 
\midrule
DDG~\cite{Liu2020DDG} & Grasping & 565 & 6.9K & GraspIt \\
DexYCB~\cite{chao2021dexycb} & Grasping &  20 & 1K & Annot. \\
ContactPose~\cite{brahmbhatt2020contactpose} & Grasping & 25 & 2306 & Capture \\
RealDex~\cite{DBLP:conf/ijcai/LiuYWWWYSYWY0M24} & Grasping &  52 & 2.6K & Capture \\
DexGraspNet~\cite{wang2023dexgraspnet} & Grasping&  5K & 1.32M & Optim. \\
SynH2R~\cite{christen2024synh2r} & Handover & 1174 & 6K & Optim. + RL \\
RP1M~\cite{zhao2024rp1m} & Piano & - &  1M & RL \\
\ourdataset\ (Ours) & Grasping + Arti. & 6K & 1B & Optim. + Gen. \\
\bottomrule
\end{tabular}
\caption{\textbf{Dataset comparison} on dexterous robotic manipulation. Our proposed \ourdataset\ is a multi-task trajectory-level dataset with 1B demonstrations. Optim., Arti., Gen. are short for optimization, articulation and generative models.}
\vspace{-15pt}
\label{table:datasets} 
\end{table}

\section{Related Work}
\label{sec:related-work}

\noindent
\textbf{Dexterous Hand Manipulation.}
Dexterous hand manipulation is a pivotal area in robotics, concentrating on precise, multi-fingered grasping and manipulation. Early research in this field primarily addressed control-based methods, laying the foundation for dexterous manipulation through studies in caging to grasping techniques \cite{rodriguez2012caging}, grasp synthesis \cite{rosales2012synthesis}, and manipulability in underactuated hands \cite{prattichizzo2012manipulability}. Further developments in control-based approaches simplified the search space for multi-fingered grasps \cite{ponce1993characterizing, ponce1997computing}, as well as optimization processes \cite{zheng2009distance, dai2018synthesis}, leading to highly precise, agile, and safe manipulation. However, these methods generally lack generalization across diverse environments and use cases.

Subsequent research shifted towards learning-based approaches to enhance flexibility and scalability \cite{andrychowicz2020learning, nagabandi2020deep}. This includes generating pose vectors directly \cite{jiang2021hand, corona2020ganhand, yang2021cpf}, utilizing intermediate representations \cite{shao2020unigrasp, wu2022learning}, and leveraging contact maps \cite{brahmbhatt2019contactgrasp, turpin2022grasp}. While learning-based methods offer increased adaptability, they are sensitive to data quality and scope, a limitation addressed in this work. Recent works~\cite{DBLP:conf/corl/LumMMHAHRW24,singh2024dextrah} employ RL-based methods to transfer robust grasping capabilities of dexterous robotic hands to the real world, using point cloud and RGB inputs, respectively.

\noindent
\textbf{Manipulation Dataset.}
Existing manipulation datasets can be broadly categorized into synthetic and real-world collections. Synthetic datasets like ObMan~\cite{hasson2019learning} and DDGdata~\cite{lundell2021ddgc} utilize the GraspIt planner to generate grasping poses but suffer from limited diversity due to naive search strategies. Real-world datasets with human hand poses offer more natural interactions, such as HO3D~\cite{hampali2020honnotate} which leverages 2D keypoint annotations and physics constraints, and DexYCB~\cite{chao2021dexycb} which captures multi-view RGBD recordings. ContactDB~\cite{brahmbhatt2019contactdb} and ContactPose~\cite{brahmbhatt2020contactpose} further enhance grasp understanding by incorporating thermal imaging and detailed contact information, respectively. However, these real-world datasets are inherently restricted to human hand structures and common daily hand poses. In contrast, our approach leverages optimization and neural networks to generate diverse manipulation trajectories that transcend these limitations.
Our work is related to prior work~\cite{lu2020multi}, which has a similar goal of expanding the distribution of generative models. Our approach differs in two ways: (1) Scale: our pipeline aims to generate 1B demonstrations rather than thousands. (2) Diversity: we increase diversity by conditioning on object geometry instead of using a classifier.
The recent work DexGraspNet2.0~\cite{zhang2024dexgraspnet} proposes learning a diffusion model over large-scale optimized demonstration datasets. In contrast, our work integrates the generative model directly into the data generation pipeline instead of relying solely on optimization. Additionally, we unify grasping and articulation tasks in both our model design and benchmark, while they focus on grasping only.
We presents the differences of several representative manipulation datasets in Tab.~\ref{table:datasets}.

\begin{figure*}
\centering
  \includegraphics[width=1.\linewidth]{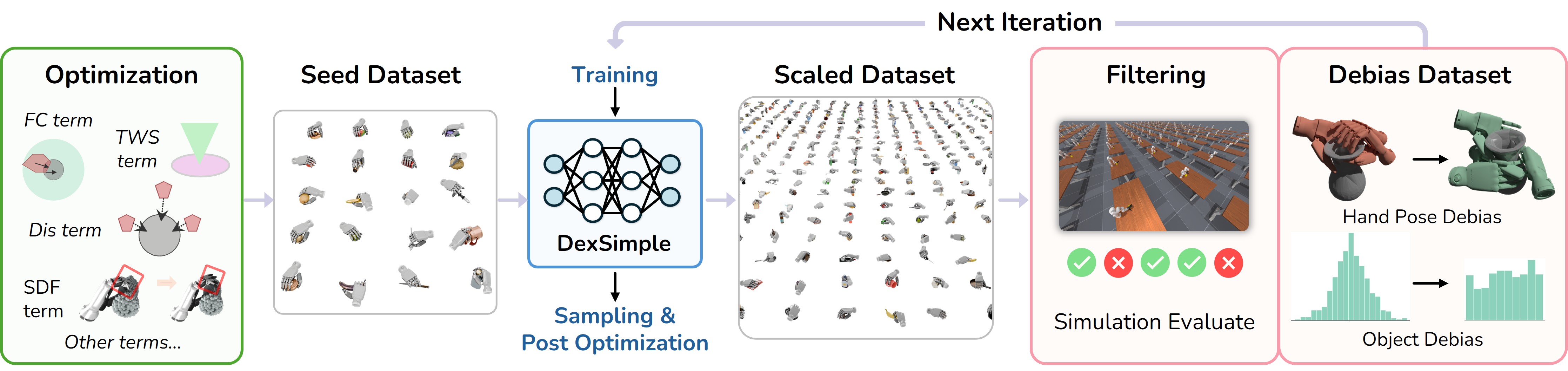}
  \vspace{-0.15in}
  \caption{\textbf{\textit{\ourdataset}} demonstration collection. The engine takes object assets and hand pose initialization as input, using a control-based optimization algorithm to generate the Seed dataset. Then the Seed dataset is used as the training data for \ourmethod, else for \ourdataset for the last iteration. Then \ourmethod~will generate a scaled proposal dataset with $\pi$ as the scaling ratio. For the proposal dataset, we then use the simulation critic and debiased algorithm to create the debiased dataset for optimization refinement.}
  \vspace{-12pt}
  \label{fig:overview}
\end{figure*}

\section{\ourdataset\ Benchmark}
\label{sec:method}

\noindent
We introduce a comprehensive benchmark for two fundamental dexterous manipulation tasks: \textbf{grasping} and \textbf{articulation}. In the grasping task, the robot hand must reach for and lift an object, whereas the articulation task requires the hand to manipulate an articulated object to achieve a specific degree of opening. Our benchmark consists of over 6,000 diverse objects and provides one billion demonstrations across three dexterous hands: the Shadow Hand, the Inspire Hand, and the Ability Hand. Each demonstration consists of a complete action sequence, from initial reaching to object manipulation.

To generate these demonstrations, we synthesize key hand poses at critical interaction points with the object, while the remaining action sequences—such as reaching, lifting, and opening—are generated using motion planning. The evaluation of our benchmark is conducted with ManiSkill~\cite{taomaniskill3, xiang2020sapien}.

\noindent\textbf{Overview of Data Generation.} 
Broadly, hand pose generation for dexterous manipulation can be approached through optimization-based methods or generative models. While optimization methods can be effective, they are often computationally expensive, especially for large-scale generation, and tend to bias the dataset toward simpler cases. On the other hand, generative models rely on an initial dataset to learn meaningful data distributions. In this work, we integrate both approaches to leverage their strengths.

As illustrated in Figure~\ref{fig:overview}, we begin by constructing a small-scale seed dataset using optimization. This seed dataset serves as the foundation for training a generative model to learn its underlying data distribution. The trained generative model is then used to produce additional demonstrations. However, since the generative model inherently inherits the biases of the seed dataset, we introduce a debiasing strategy to enhance diversity. Specifically, we condition the generative model on targeted factors to generate hand poses under less frequently observed conditions, thereby expanding the dataset beyond the initial distribution. By iteratively refining the generative model through repeated training and debiasing operations, we construct our final dataset, \ourdataset, which achieves both diversity and robustness in dexterous manipulation demonstrations.

\noindent
\textbf{Optimization for Seed Dataset.} To generate the seed dataset, we implement an efficient optimization method for hand pose synthesis based on previous work~\cite{wang2023dexgraspnet,chen2023task}, while including new features like scene-level collision avoidance and support for various hands. Although the optimization process is well-engineered (1,000 grasps per minutes on a single GPU), generating one billion demonstrations remains computationally expensive. Therefore, we only use optimization to create a small-scale seed dataset (around 5 million poses).
A dexterous hand pose is parameterized by a tuple \(g = (T, R, \theta)\), where \(T \in \mathbb{R}^3\) for global translation, \(R \in SO(3)\) for global rotation, and \(\theta \in \mathbb{R}^d\) for robot hand joint angles (d = 22 for Shadow hand, d = 6 for Inspire and Ability hand). The object geometry is represented by mesh \(O\). Unlike previous methods~\cite{wang2023dexgraspnet,chen2023task}, which use link meshes to model hand geometry, we approximate the hand using manually defined spheres (around 10 spheres for each link). This approximation significantly accelerates the optimization process. 

Our optimization energy function for the grasping task \( E_{\texttt{grasp}} \) is given by:  
\[
E_{\texttt{grasp}} = E_{\texttt{fc}} + w_{\texttt{dis}}E_{\texttt{dis}} +  w_{\texttt{sdf}}E_{\texttt{sdf}} + w_{\texttt{j}}E_{\texttt{j}} + w_{\texttt{s}}E_{\texttt{s}},
\]
where \( E_{\texttt{fc}} \) is the force closure energy term, \( E_{\texttt{dis}} \) minimizes contact distance, \( E_{\texttt{sdf}} \) prevents penetration, \( E_{\texttt{j}} \) enforces joint limits, and \( E_{\texttt{s}} \) avoids self-collisions. \( w_{x} \) represents the weight for the corresponding terms. The penetration term \( E_{\texttt{sdf}} \) is formulated based on the sphere-based hand representation. A sphere is defined by \( (c, r) \), where the center \( c \in \mathbb{R}^3 \) and the radius \( r \in \mathbb{R}^+ \). The center \( c \) is transformed with the link pose using forward kinematics. The SDF penetration term is given by:
\[
E_{\texttt{sdf}} = \sum_i r_i - SDF(c_i, O),
\]
where \( SDF(\cdot) \) is the signed distance function (SDF) query between a 3D point and a mesh. To prevent scene collisions, we incorporate SDF queries for other meshes (e.g., the table), and the final SDF for a sphere is computed as the maximum value across different meshes. Detailed formulation of other energy terms can be found in previous works~\cite{wang2023dexgraspnet,liu2021synthesizing}.

While the force closure energy term \( E_{\texttt{fc}} \) is suitable for the grasping task, achieving force closure in the articulation task is usually difficult and unnecessary. Instead, the articulation task requires the hand pose to generate specific forces and torques, such as rotating the top part of a laptop or pulling a drawer. Therefore, we replace the force closure energy term \( E_{\texttt{fc}} \) with the task wrench space term \( E_{\texttt{tws}} \) introduced in ~\cite{chen2023task}, which approximates the difference between a target wrench space and the current wrench space of a given hand pose. The optimization energy function for the articulation task \( E_{\texttt{arti}} \) is defined as:
\[
E_{\texttt{arti}} = E_{\texttt{tws}} + w_{\texttt{dis}}E_{\texttt{dis}} +  w_{\texttt{sdf}}E_{\texttt{sdf}} + w_{\texttt{j}}E_{\texttt{j}} + w_{\texttt{s}}E_{\texttt{s}}.
\]
The target wrench space for articulated objects with revolute joints consists of a torque aligned with the joint axis and arbitrary forces. The wrench space for objects with prismatic joints consists of forces sampled from a \(30^\circ\) cone aligned with the joint axis and zero torque.

The optimization process is implemented using Warp-Lang~\cite{warp2022} and accelerated with its BVH mesh structure. The optimized hand poses are evaluated using the simulator, and the successful ones are retained as a seed dataset.

\begin{figure*}
\centering
  \includegraphics[width=1.\linewidth]{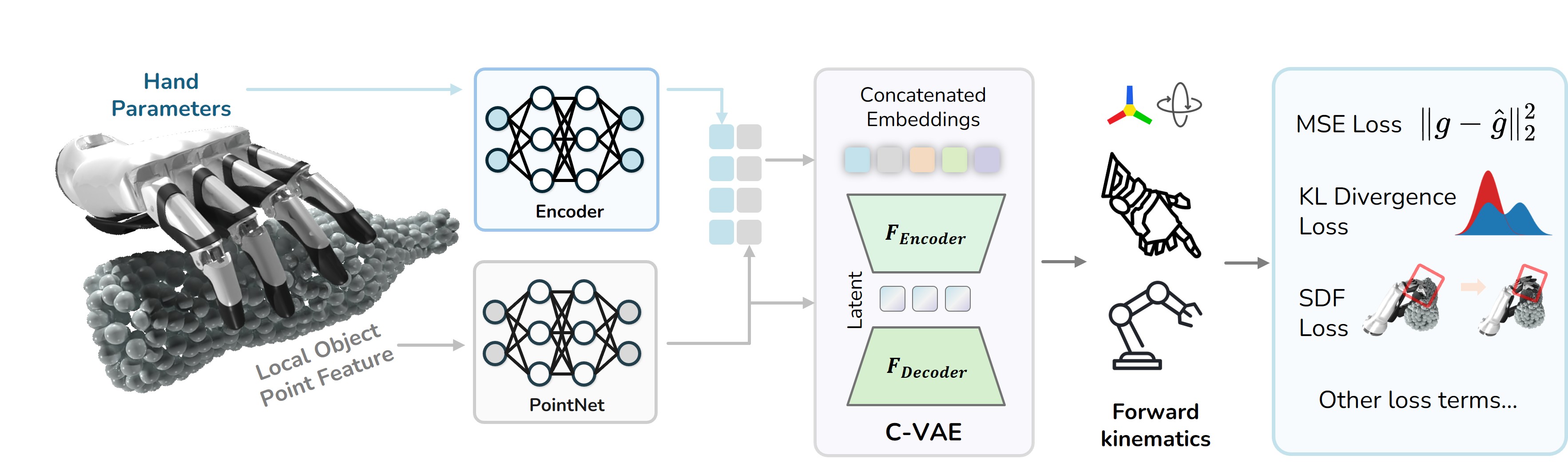}
  \vspace{-10pt}
  \caption{\textbf{\textit{\ourmethod~Pipeline}}. Our model takes in hand parameters and object point clouds as fixed input for CVAE, while root rotation, translation, and joint value as optional conditions. Those are combined as the input embeddings for CVAE, while point cloud embeddings are re-emphasized at the latent space. The output of CVAE is the forward kinematics define the hand pose trajectory optimized by the effective loss function.}
  \label{fig:architecture}
  \vspace{-10pt}
\end{figure*}

\noindent
\textbf{Generative Models for Scaling-up Demonstrations.}
Generative models are widely adopted for capturing the distribution of action demonstrations. However, applying these models for data generation still presents several challenges: i). \textbf{Feasibility:} The success rate of generative models is often lower than that of deterministic models, leading to a higher proportion of infeasible samples. ii). \textbf{Limited Diversity:} While generative models can produce more diverse actions than deterministic models, they still tend to interpolate between the demonstrations, which may maintain or even reduce the original level of diversity of whole dataset rather than expanding it.

To address the feasibility issue, we first incorporate geometric constraints during the generation process, enabling our model to outperform state-of-the-art generative models (see Sec.~\ref{subsec:model} for model details and Sec.~\ref{sec:exp} for experiments). In addition, we apply a post-optimization step to the sampled hand poses to prevent penetration and ensure that the fingers closely cover the object. The energy function for the post-optimization stage of both tasks, \( E_{\texttt{post}} \), is defined as:
\[
E_{\texttt{post}} = w_{\texttt{dis}}E_{\texttt{dis}} +  w_{\texttt{sdf}}E_{\texttt{sdf}} + w_{\texttt{j}}E_{\texttt{j}} + w_{\texttt{s}}E_{\texttt{s}}.
\]
We exclude the task-specific terms \( E_{\texttt{fc}} \) and \( E_{\texttt{tws}} \) here since we only aim to make slight adjustments to the finger positions.  Although we use optimization in this stage, the overall data generation, combined with generative models, remains significantly more efficient than pure optimization. The sampling process is approximately 100 times faster, and the number of iterations in post-optimization is substantially lower than in the pure optimization (100 vs. 6000). The refined hand poses are then evaluated using the simulator. 

To improve diversity, we encourage the generative model to sample actions that differ more from existing actions in the dataset while maintaining success rate. To achieve this, we introduce an additional condition to the generative model and prioritize sampling actions for less frequent conditions. Specifically, we associate each hand pose with a single 3D point on the object. We first define the heading direction \(v \in \mathbb{R}^3\) of a hand pose as the vector from the palm center to the midpoint between the thumb tip and the middle finger tip. The closest point along this direction is then assigned as the associated point of the hand pose. We adapt our generative model to take the feature vector of a 3D point as a condition for generating corresponding actions. During data generation, we first statistically compute the probability of each point associated with existing actions on the object and then sample new actions inversely proportional to this probability. Additionally, we statistically count the number of existing actions for each object and sample more actions for the more challenging ones.

After increasing the dataset size and diversity, retraining the model on the expanded dataset can further improve its performance. This \textit{iterative data generation} process can be repeated multiple times to progressively refine both the model and the dataset.

\noindent
\textbf{Motion Planning.} With the key-frame action, motion planning is still required to complete both tasks. For the reaching stage in both tasks, motion planning can be formulated as an optimization problem that maximizes smoothness while avoiding collisions.  Given a manually defined starting action and a goal action, we first linearly interpolate between them, and then optimize the intermediate actions using the following energy function:
\[
E_{\texttt{reach}} = w_{\texttt{smooth}} \sum_{i=1}^{N} \|g_i - g_{i-1}\|^2 + w_{\texttt{sdf}} \sum_{i=0}^{N} E_{\texttt{sdf}}(g_i),
\]
where \( \{g_0, g_1, \dots, g_N\} \) denotes the sequence of hand poses along the trajectory, \( w_{\texttt{smooth}} \) and \( w_{\texttt{sdf}} \) are weights for smoothness and collision avoidance respectively, and \( E_{\texttt{sdf}}(g_i) \) is computed based on the sphere-based SDF errors with respect to nearby scene meshes. Minimizing \( E_{\texttt{reach}} \) produces a trajectory that is both smooth and collision-free. Compared to other motion planning libraries, this simple optimization is naturally suitable for large-scale parallel data generation.

After reaching, for the grasping task, we will execute an over-shoot action to grasp the object and increase the height of the target action to lift it. For the articulation task, we will follow the trajectory based on the given joint axis (rotate along the revolute joint axis or translate along the prismatic joint axis). The complete planned trajectory is executed in the simulator for evaluation.

\section{\ourmethod\ Model}
\label{subsec:model}

While a large body of generative models~\cite{jiang2021graspTTA,lu2023ugg,ye2023learning,liu2023contactgen,weng2024DexDiffuser} have been proposed for dexterous hand manipulation in recent years, their use for data generation or policy deployment remains limited. In this work, we revisit the simple CVAE model and demonstrate that incorporating an SDF-based geometric constraint during training enables it to outperform state-of-the-art methods by a large margin. Furthermore, we integrate additional condition over the base model to support diverse data generation.

\noindent
\textbf{Vision Encoder and CVAE.} We employ a point cloud \(P \in \mathbb{R}^{N\times3}\) as the visual input, using a full point cloud sampled from the object mesh for data generation and a single-view depth map for policy deployment. We utilize PointNet~\cite{qi2017pointnet} to encode the point cloud into a global feature vector \( f_{\texttt{obj}} \in \mathbb{R}^{d}  \) and local feature vectors \( f_{p} \in \mathbb{R}^{d} \) for each point \( p \in \mathbb{R}^{3} \):
$$
f_{\texttt{obj}}, \{ f_{p} \}_{p \in P} = \textbf{PointNet}(P).
$$
The VAE model uses a multi-layer perceptron (MLP) to encode the hand pose $g$ into the mean and standard deviation vectors of a latent distribution. A sample is drawn from this distribution and passed to the MLP decoder to reconstruct the original hand pose. After concatenating conditional vectors (e.g., the global point cloud feature vector \( f_{\texttt{obj}} \)) to both the inputs of the VAE encoder and decoder, the CVAE model can generate samples under a given condition:
\begin{align}
    \mu, \sigma &= \textbf{Enc}(g, f_{\texttt{obj}}), \nonumber \\
    z &= \mu + \sigma \odot \epsilon, \quad \epsilon \sim \mathcal{N}(0, I), \nonumber \\
    \hat{g} &= \textbf{Dec}(z, f_{\texttt{obj}}).
\end{align}
In our work, we simply concatenate additional vectors to incorporate more conditions.

\noindent
\textbf{Objectives.} The CVAE training is supervised by the standard reconstruction loss $\mathcal{L}_\texttt{R}  = \| g - \hat{g} \|_2^2$ and the KL divergence loss $\mathcal{L}_\texttt{KL} = D_{\mathrm{KL}}\Big( \mathcal{N}(\mu, \sigma^2) \,\Big\|\, \mathcal{N}(0, I) \Big)$. To enforce geometric constraints, we introduce an SDF-based loss. Different from the optimization stage, building a BVH structure for each object at each iteration during training is time-consuming. We therefore use a sampled point cloud to represent the object geometry while still using spheres to represent the hand geometry. The SDF loss is then given by:
\[
\mathcal{L}_\texttt{SDF} = \sum_{c \in \mathcal{C}} \max\Big(0,\; r_c - \min_{p \in P} \| c - p \|_2 \Big)^2,
\]
where \( \mathcal{C} \) is the set of spheres representing the hand geometry and \( r_c \) is the radius of sphere \( c \). This loss is not only straightforward to implement in PyTorch, but we also empirically find that training with the point-sphere SDF loss is more stable compared to the mesh-sphere SDF loss used in the optimization stage. With this SDF loss term, our proposed \ourmethod\ outperforms SOTA methods by a large margin (see Sec.\ref{sec:exp}).

The overall loss for the base model is defined as:
\[
\mathcal{L} = \lambda_\texttt{R} \mathcal{L}_\texttt{R} + \lambda_\texttt{KL} \mathcal{L}_\texttt{KL} + \lambda_\texttt{SDF} \mathcal{L}_\texttt{SDF},
\]
where \( \lambda_\texttt{R} \), \( \lambda_\texttt{KL} \), and \( \lambda_\texttt{SDF} \) are weights that balance different loss terms.

\noindent
\textbf{Conditions for Data Generation.} As mentioned in Sec.~\ref{sec:method}, each hand pose is associated with a single object 3D point \(p\) by finding the closest point along its heading direction \(v\). To achieve this, we condition the CVAE on the corresponding local object feature vector \(f_p\).

\section{Experiments}
\label{sec:exp}
We firstly evaluate the effectiveness of the proposed generative model, \ourmethod, for grasp synthesis on the DexGraspNet~\cite{wang2023dexgraspnet} benchmark. Then, we provide details on the synthesized \ourdataset\ demonstration dataset and compare it with human-annotated demonstration datasets on both lifting and articulation tasks. Additionally, we downscale \ourdataset\ and evalute the performance of methods trained on it. Finally, ablation studies are conducted to validate our design choices.

\subsection{Grasping Synthesis Evaluation}
\label{subsec:grasping}

Grasping is essential in most manipulation tasks, we firstly evalute the proposed method's effectiveness in grasp synthesis using the DexGraspNet~\cite{wang2023dexgraspnet} benchmark. We train \ourmethod\ solely with the benchmark’s provided training data, reducing the output to a single frame and omitting conditioning during training.
The validation is conducted in the Isaac Gym simulator~\cite{makoviychuk2021isaac} using ShadowHand~\cite{shadowhand2005}. We adhere to the metrics established in the benchmark to ensure fair comparisons with baseline methods, which are divided into two categories: quality (Success Rate, $Q_1$-score, Penetration) and diversity (H mean and H std). We follow the implementations \footnote{Please refer to supp. material for details.} in ~\cite{wang2023dexgraspnet, lu2023ugg}. We compare with DDG~\cite{Liu2020DDG}, GraspTTA~\cite{jiang2021graspTTA}, the generation module in UniDexGrasp~\cite{xu2023unidexgrasp} (abbreviated as UDG), and UGG~\cite{lu2023ugg}.

We present quantitative results in Table~\ref{tab:grasp}.
Many grasp generation methods, such as UGG~\cite{lu2023ugg}, commonly employ post-optimization to enhance performance. To ensure a fair comparison, we indicate the use of post-optimization (abbreviated as ``Opt'') in the table.  The results show that the proposed generative model, \ourmethod, outperforms all baseline methods by a large margin. In terms of quality, \ourmethod\ (with post-optimization) achieves the highest success rate ($86.0\%$), the highest $Q_1$ score ($0.125$), and the lowest penetration ($0.13$). For diversity, \ourmethod\ outperforms baseline with a higher entropy mean of $8.56$.

\begin{table}[t]
\small
\centering
\setlength{\tabcolsep}{2pt}
\fontsize{8pt}{9pt}\selectfont
\begin{tabular}{lcclccclcc}
\toprule
& \multicolumn{2}{c}{Setting} & & \multicolumn{3}{c}{Quality} & &  \multicolumn{2}{c}{Diversity} \\
\cmidrule(lr){2-3} \cmidrule(lr){5-7} \cmidrule(lr){9-10}
Method & Opt & Filter & & SR $\uparrow$ & $Q_1$ $\uparrow$ & Pen $\downarrow$ & & H mean $\uparrow$ & H std $\downarrow$ \\
\midrule
DDG~\cite{Liu2020DDG} & & & & 67.5 & 0.058 & 0.17 & & 5.68 & 1.99 \\
UGG~\cite{lu2023ugg} & & & & 43.6 & 0.026 & 0.43 & & 8.33 & 0.30 \\
\ourmethod & & & & 63.7 & 0.075 & 0.29 & & 8.53 & 0.25  \\
\midrule
UDG~\cite{xu2023unidexgrasp} & $\checkmark$ & & & 23.3 & 0.056 & 0.15 & & 6.89 & 0.08 \\
GraspTTA~\cite{jiang2021graspTTA} & $\checkmark$ & & & 24.5 & 0.027 & 0.68 &  & 6.11 & 0.56 \\
UGG~\cite{lu2023ugg} & $\checkmark$ & & & 64.1 & 0.036 & 0.17 & & 8.31 & 0.28 \\
\rowcolor{myblue}
\ourmethod & $\checkmark$ & & & \textbf{86.0} & \textbf{0.125} & \textbf{0.13} & & \textbf{8.56} & 0.15 \\
\midrule
\gc{UGG~\cite{lu2023ugg}} & \gc{$\checkmark$} & \gc{$\checkmark$} & & \gc{72.7} & \gc{0.063} & \gc{0.14} & & \gc{7.17} & \gc{0.07} \\
\gc{\ourmethod} & \gc{$\checkmark$} & \gc{$\checkmark$} & & \gc{92.6} & \gc{0.132} & \gc{0.12} & & \gc{8.56} & \gc{0.16} \\
\bottomrule
\end{tabular}
\caption{\textbf{Grasping synthesis} results on the DexGraspNet~\cite{wang2023dexgraspnet} benchmark. The proposed generative model, \ourmethod, significantly outperforms all baseline methods. Some evaluation results are taken from UGG~\cite{lu2023ugg}. Opt, SR, and Pen are short for Optimization, Success Rate, and Penetration, respectively.}
\vspace{-2em}
\label{tab:grasp}
\end{table}

\begin{figure*}[t]
\centering
  \includegraphics[width=1.0\linewidth]{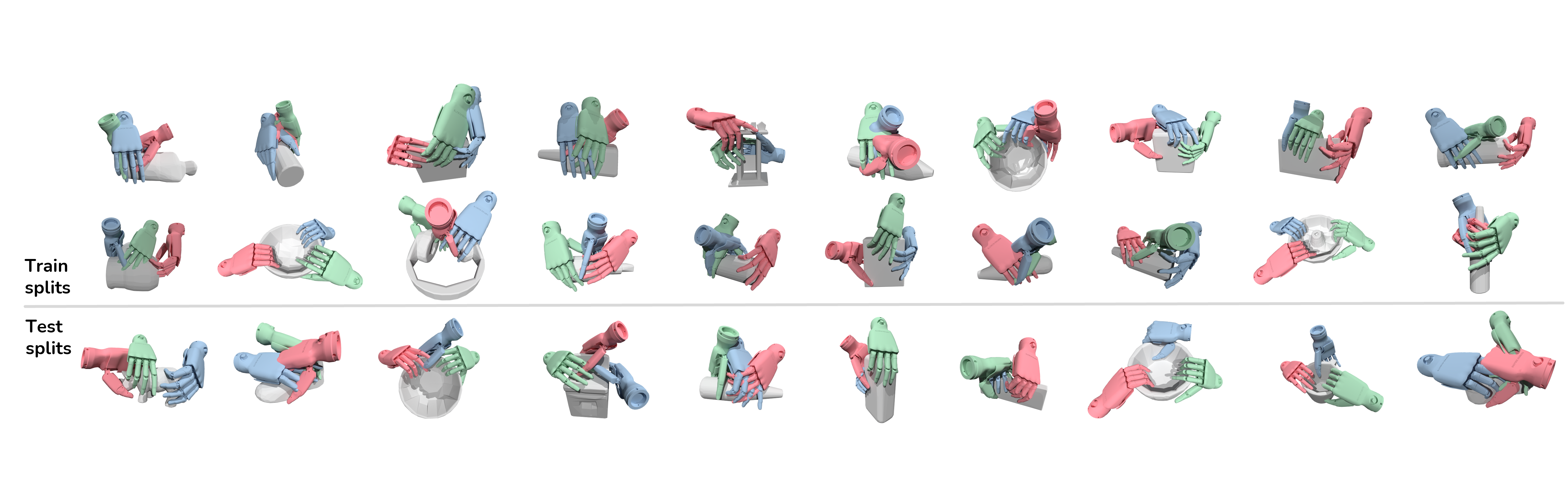} 
  \vspace{-3em}
  \caption{\textbf{Diverse demonstrations} for objects from train/test splits. We show only the contact frame for clarity.}
  \vspace{-1em}
  \label{fig:diversity}
\end{figure*}

UGG~\cite{lu2023ugg} proposes a learning-based discriminator to filter grasping, which can be applied to our method. With this filtering, the success rate increases to $92.6\%$.  It is worth noting the success rate of \ourmethod\ without post-optimization and filtering is slightly lower than that of DDG~\cite{Liu2020DDG}; this is expected as our method is a generative model while DDG~\cite{Liu2020DDG} is a regression model, and our method achieves much higher diversity ($8.53$ vs. $5.68$). 

\begin{figure}[t!]
\centering
  \vspace{-20pt}
  \includegraphics[width=0.9\linewidth]{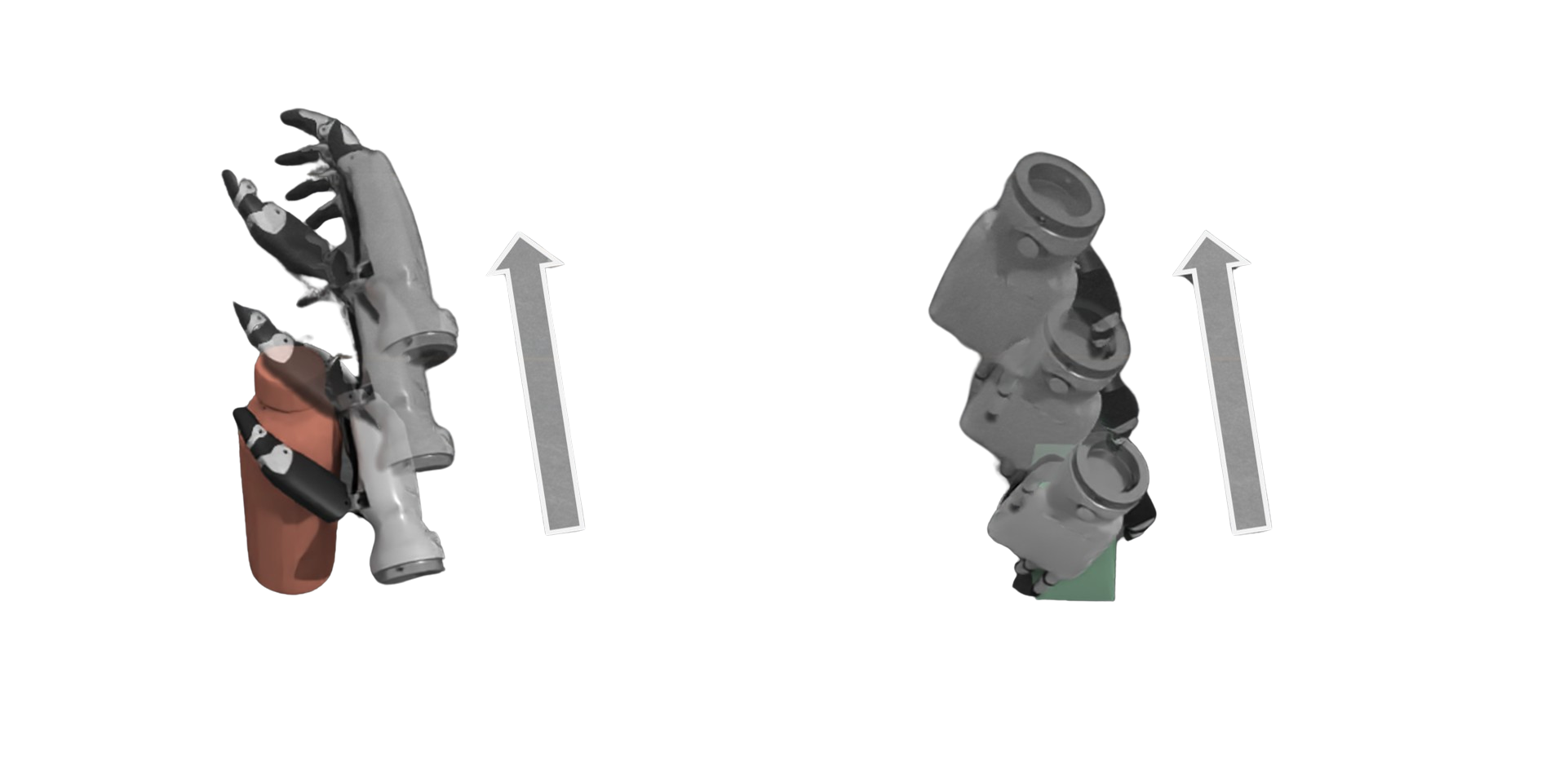} 
  \vspace{-25pt}
  \caption{\textbf{Lifting trajectory} from \ourdataset\ dataset.}
  \vspace{-1em}
  \label{fig:lifting}
\end{figure}

\subsection{Dataset Analysis}

\noindent
\textbf{Tasks Definition.} While the proposed iterative data generation pipeline can be applied to multiple hands, we take ShadowHand as an example to detail our data curation process. Beyond the grasping synthesis task in Sec.~\ref{subsec:grasping}, we focus on two tasks: \textbf{Grasping} and 
\textbf{Articulation}. The goal of the grasping task is to reach an object placed on a table, grasp it, and lift it to a specified height (0.4 m) while maintaining contact between the hand and the object. We show example trajectories from \ourdataset\ for the grasping task in Fig.~\ref{fig:lifting}. The articulation task requires reaching an articulated structure (e.g., a laptop, box, or faucet) and opening it to increase its joint angle by 0.5 while maintaining contact between the hand and the object.

For the grasping task, we utilize all 5751 object assets collected by DexGraspNet~\cite{wang2023dexgraspnet} and exclude all objects that cannot stand stably on the table. Note that the settings between DexGraspNet and our dataset are different. Our focus is on table-top tasks, while DexGraspNet focuses on grasping objects in free space. Approximately $90\%$ of grasping demonstration in DexGraspNet would collide with the table.  For the articulation task, we utilize 650+ articulation assets collected by DexArt~\cite{bao2023dexart} and ManiSkill~\cite{taomaniskill3}. We adopt the official training/testing splits provided by previous works~\cite{wang2023dexgraspnet,fan2023arctic,bao2023dexart}.

\noindent
\textbf{Dataset Curation.}  Our dataset curation starts by optimizing a seed dataset with 5 million demonstrations. We first train \ourmethod\ on this seed dataset to create a 50 million proposal dataset. After optimization refinement, we use the ManiSkill~\cite{taomaniskill3}/SAPIEN~\cite{xiang2020sapien} simulation to filter out unsuccessful trajectories and rebalance the data as described in Sec.~\ref{sec:method}. We then retrain \ourmethod\ on the 50 million dataset to produce a 500 million proposal dataset, and repeat this process. Finally, we collect a dataset with 950 million (around 1 billion) successful trajectories.

\begin{figure}[t] 
\centering
  \includegraphics[width=1.0\linewidth]{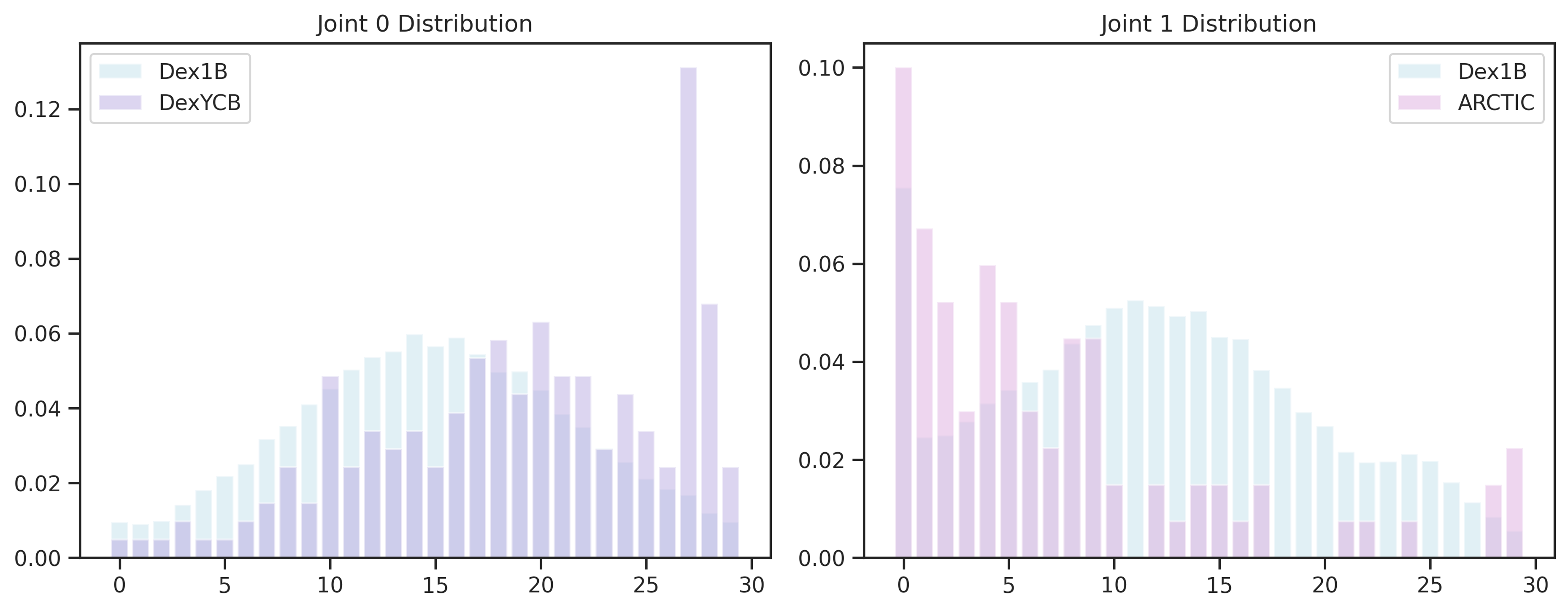} 
  \caption{\textbf{Probability distribution of joint values} from \ourdataset\ and DexYCB/ARCTIC. The distribution of \ourdataset\ is more evenly spread, centering around the mean joint values.}
  \vspace{-1.3em}
  \label{fig:diversity-comparison}
\end{figure}

\noindent
\textbf{Implementation Details.} Compared to DexGraspNet~\cite{wang2023dexgraspnet}, our implementation of pure optimization-based grasp generation is 30 times faster, requiring only 2 minutes to generate 2000 grasps for 6000 steps on a single RTX-3090, while also achieving a higher success rate ($27\%$ vs. $20\%$ for ShadowHand). When initialized with the network, our method is even 700 times faster, including CVAE sampling and 100 post-optimization steps. All CUDA and geometry-related operations are implemented using warp-lang~\cite{warp2022}.

In the proposed \ourmethod, we adopt PointNet~\cite{qi2017pointnet} as the visual encoder, extracting object features in a dimension of 256. For the CVAE, both the input and output dimensions are $N_{\text{frame}} \times N_{\text{DOF}}$, and the latent vector dimension is set to 256.

\begin{figure*}[t]
\centering
  \includegraphics[width=1.0\linewidth]{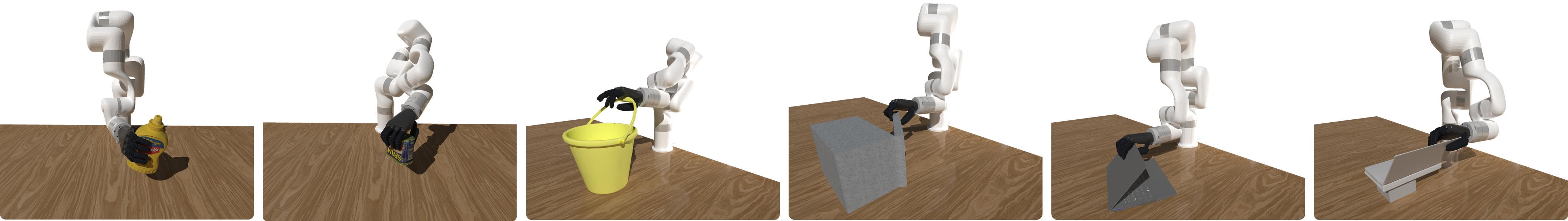} 
  \caption{\textbf{Qualitative results} for both grasping and articulation tasks. We show only the contact frame for clarity.}
  \vspace{-1em}
  \label{fig:viz}
\end{figure*}

\noindent
\textbf{Dataset Comparison}. We demonstrate the quality of \ourdataset\ by comparing it to two large-scale, human-annotated, trajectory-level datasets: DexYCB~\cite{chao2021dexycb} and ARCTIC~\cite{fan2023arctic}. DexYCB includes 20 objects, with approximately 500 trajectories for each hand. ARCTIC includes 10 articualtion objects with a total of 
301 trajectories. We follow ~\cite{ye2023learning,chen2022dextransfer} to generate robot demonstrations from the DexYCB and ARCTIC dataset, including retargeting human demonstrations to robot trajectories and adding noise to generate a larger number of physically plausible demonstrations. We collected $62\%$ and $64\%$ of all trajectories from the DexYCB and ARCTIC datasets, respectively, that successfully achieve task goals.

We highlight the diversity of \ourdataset. Using the proposed debiasing approach in Sec.\ref{sec:method}, the diversity across object categories, joint values, and wrist poses in our dataset can be easily enhanced by generating additional samples for underrepresented data. Besides, we discretize the range of joint angles into bins and estimate a probability distribution over them. Fig.~\ref{fig:diversity-comparison} shows the distribution of two joints. Unlike DexYCB and ARCTIC, which often have joint values concetrate at the limits, the distribution of \ourdataset\ is more evenly spread, centering around the mean joint values. This is achieved by debiasing approach and including a regularization term that discourages the hand from getting too close to the joint limits during optimization. Qualitative results are presented in Fig.~\ref{fig:diversity}.

\begin{table}[t] 
\small
\centering
\setlength{\tabcolsep}{2pt}
\fontsize{8pt}{9pt}\selectfont
\begin{subtable}[t]{0.95\linewidth}
\centering
\begin{tabular}{lccccc}
\toprule
& & \multicolumn{2}{c}{{\scriptsize Eval on} DexYCB} & \multicolumn{2}{c}{{\scriptsize Eval on} \ourdataset} \\
\cmidrule(lr){3-4} \cmidrule(lr){5-6} 
Method & Training Data & Train set & Test set & Train set & Test set \\
\midrule
BC w. PointNet & DexYCB~\cite{chao2021dexycb} & 34.72 & 3.03 & 1.02 & 2.56 \\
\ourmethod & DexYCB~\cite{chao2021dexycb} & 43.49 & 21.21 & 23.68 & 22.80 \\
\midrule
BC w. PointNet & \ourdataset\ (ours) & 33.02 & 31.82 & 31.40 & 28.54 \\
\rowcolor{myblue}
\ourmethod & \ourdataset\ (ours) & \textbf{47.17} & \textbf{53.02} & \textbf{49.58} & \textbf{45.40} \\
\bottomrule
\end{tabular}
\caption{Lifting task comparison on DexYCB~\cite{chao2021dexycb} and \ourdataset.}
\end{subtable}
\hfill
\begin{subtable}[t]{0.95\linewidth}
\centering
\begin{tabular}{lccccc}
\toprule
& & \multicolumn{2}{c}{{\scriptsize Eval on} ARCTIC} & \multicolumn{2}{c}{{\scriptsize Eval on} \ourdataset} \\
\cmidrule(lr){3-4} \cmidrule(lr){5-6} 
Method & Training Data & Train set & Test set & Train set & Test set \\
\midrule
BC w. PointNet & ARCTIC~\cite{fan2023arctic} & 41.03 & 25.62 & 37.65 & 30.16 \\
\ourmethod & ARCTIC~\cite{fan2023arctic} & 48.75 & 23.08 & 49.16 & 51.57 \\
\midrule
BC w. PointNet & \ourdataset\ (ours) & 57.50 & 63.67 & 64.74 & 56.88 \\
\rowcolor{myblue}
\ourmethod & \ourdataset\ (ours) & \textbf{72.00} & \textbf{73.49} & \textbf{77.05} & \textbf{64.79} \\
\bottomrule
\end{tabular}
\caption{Articulation task comparison on ARCTIC~\cite{fan2023arctic} and \ourdataset.}
\end{subtable}
\caption{Benchmarks on \textbf{(a) lifting tasks} with DexYCB~\cite{chao2021dexycb} and our datasets, and \textbf{(b) articulation tasks} with ARCTIC~\cite{fan2023arctic} and our datasets. Models trained on \ourdataset\ consistently outperform those trained on DexYCB/ARCTIC across various tasks, baselines, and splits.}
\vspace{-1.7em}
\label{tab:benchmark}
\end{table}

\noindent
\textbf{Benchmarks}.
We benchmark two methods for grasping and articuation tasks on our datasets, and compare them with the same methods trained on DexYCB~\cite{chao2021dexycb} and ARCTIC~\cite{fan2023arctic}. In addition to the proposed \ourmethod, we implement a vanilla behavioral cloning with PointNet~\cite{qi2017pointnet} (referred to as BC w. PointNet). This model takes the object point cloud, current hand joint values, and poses as input to predict chunked actions for the next $n=50$ frames. The predicted actions are then merged using a temporal weighting technique form ACT~\cite{DBLP:conf/rss/ZhaoKLF23}.

The results are reported in Tab.~\ref{tab:benchmark}. 
When comparing models trained on \ourdataset\ to those trained on DexYCB/ARCTIC, we consistently find that the former outperforms the latter across tasks, baselines and splits. This suggests that supervised learning methods perform better when trained on our larger and more diverse \ourdataset\ dataset. Tab.~\ref{tab:benchmark} also demonstrates that the proposed generative method, \ourmethod, achieves better performance than the regression-based BC baselines on both the relatively small DexYCB/ARCTIC dataset and the larger-scale \ourdataset. For lifting task, it also can be clearly observed that models trained on DexYCB struggle to generalize to unseen objects. Qualitative results are shown in Fig.~\ref{fig:viz}.

\begin{figure}[t]
\centering
  \includegraphics[width=1.0\linewidth]{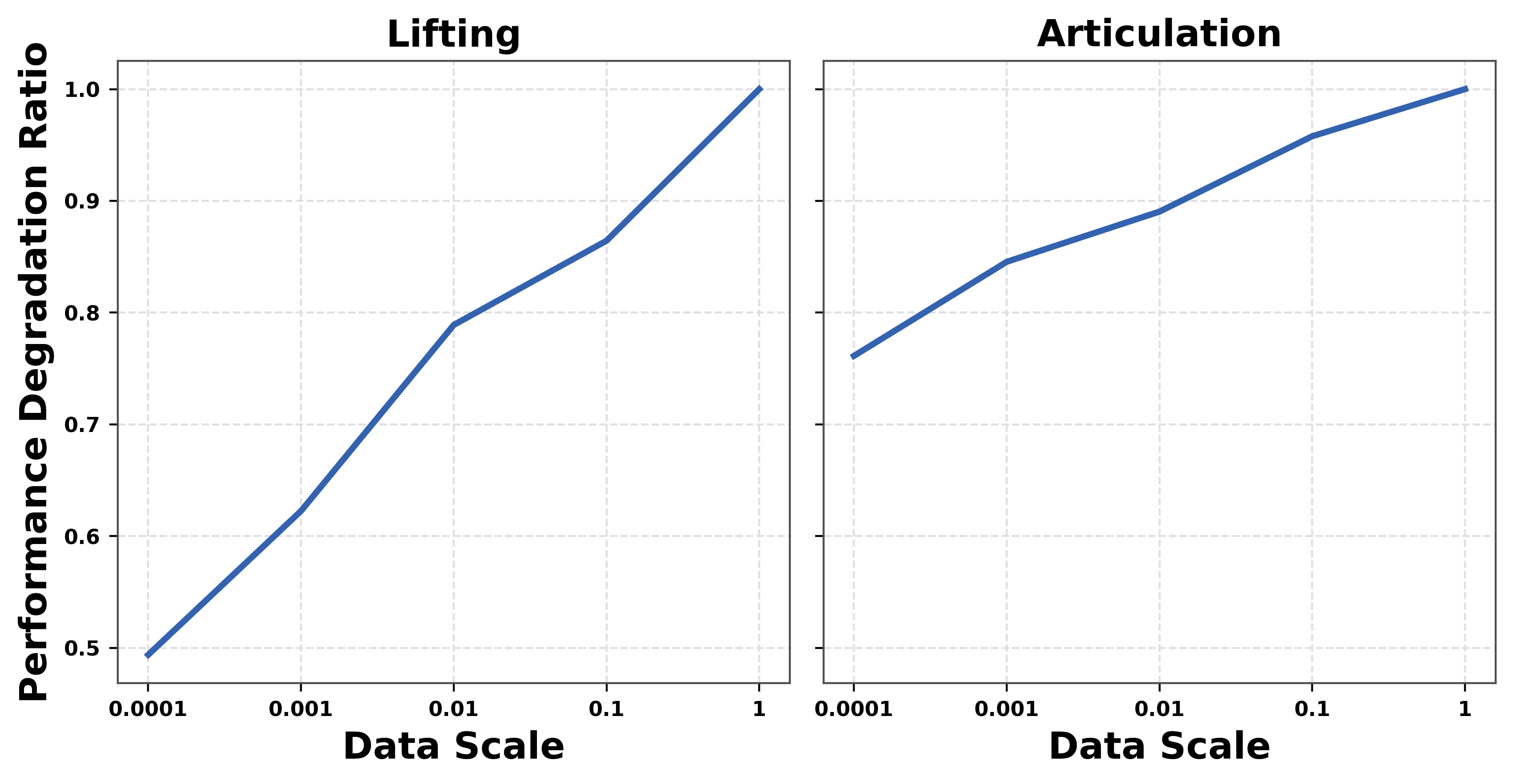} 
  \vspace{-1em}
  \caption{\textbf{Scaling the number} of demonstrations used for training. For both tasks, our model consistently improves with more training data.}
  \vspace{-2em}
  \label{fig:scaling-law}
\end{figure}

\subsection{Scaling the Dataset}
To investigate the effect of training data size on performance, we reduce the amount of training data and analyze its impact on the success rates of both the lifting and articulation tasks. As shown in Fig.~\ref{fig:scaling-law}, the performance degradation ratio increases as data is reduced, illustrating that the success rates of the proposed \ourmethod\ consistently improve with more training data. Notably, we observe that performance degradation is more pronounced for the lifting task than for the articulation task as training data decreases. We hypothesize this is because lifting relies heavily on stable object grasping, requiring a precise geometric understanding of individual objects, which becomes more challenging with reduced data. In contrast, the articulation task, which emphasizes trajectory execution, shows greater resilience to data reduction as it can adapt to unseen objects through a more generalized approach to motion. This suggests that while both tasks benefit from larger datasets, lifting requires a more extensive dataset to achieve stable performance, whereas articulation maintains reasonable performance even with less data.

\begin{table}[t]
\small
\centering
\setlength{\tabcolsep}{2pt}
\fontsize{8pt}{9pt}\selectfont
\begin{tabular}{l l ccc l cc}
\toprule
& & \multicolumn{3}{c}{Quality} & &  \multicolumn{2}{c}{Diversity} \\
\cmidrule(lr){3-5} \cmidrule(lr){7-8}
Method & & Success Rate $\uparrow$ & $Q_1$ $\uparrow$ & Penetration $\downarrow$ & & H mean $\uparrow$ & H std $\downarrow$ \\ 
\midrule
w/o. $\mathcal{L}_{\texttt{sdf}}$ & & 0.7 & 0.001 & 0.92 &  & 8.58 & 0.16 \\
w/o. $\mathcal{L}_{\texttt{D}}$ & & 42.0 & 0.044 & 0.23 & & 8.65 & 0.16 \\
Full Model & &  63.7 & 0.075 & 0.29 & & 8.53 & 0.25 \\
\bottomrule
\end{tabular}
\vspace{-0.5em}
\caption{\textbf{Ablation Study} of the geometric loss terms in grasp synthesis. Both $\mathcal{L}{\texttt{sdf}}$ and $\mathcal{L}{\texttt{D}}$ are crucial for the grasping quality.}
\vspace{-10pt}
\label{tab:ablation} 
\end{table}

\subsection{Ablation Study}
In this section, we ablate the geometric loss terms of the \ourmethod\ generative model using the DexGraspNet benchmark, with results detailed in Tab.~\ref{tab:ablation}. The \textit{sphere-representation SDF loss}, denoted as $\mathcal{L}{\texttt{sdf}}$, is designed to provide fine-grained geometric guidance, which plays a crucial role in preventing the model from penetrating the object during grasp synthesis. This loss term is essential for grasping quality, as \textit{removing $\mathcal{L}{\texttt{sdf}}$ causes a drastic drop in success rate from 63.7 to 0.7}. Without $\mathcal{L}{\texttt{sdf}}$, the model lacks precise spatial awareness, leading to significant failures in grasp execution. On the other hand, the \textit{distance loss $\mathcal{L}{\texttt{D}}$} is responsible for encouraging the hand to make stable contact with the object surface, which enhances the grasp’s stability. This loss has a notable impact on both the success rate and the $Q_1$ quality metric. Although \textit{$\mathcal{L}{\texttt{D}}$ slightly increases the penetration value}, it significantly contributes to an improved success rate and $Q_1$ score, highlighting its importance in achieving reliable grasps. The diversity metrics, represented by H mean and H std, are only minimally impacted by both loss terms, indicating that these geometric losses focus more on grasp quality than diversity. In summary, \textit{both $\mathcal{L}{\texttt{sdf}}$ and $\mathcal{L}_{\texttt{D}}$ are indispensable for high-quality grasp synthesis}, as they address different aspects of the grasping process—object penetration prevention and stable contact establishment, respectively.

\section{Real-world Experiments}
\label{sec:demo}

\begin{figure}[t]
\centering
  \includegraphics[width=1.0\linewidth]{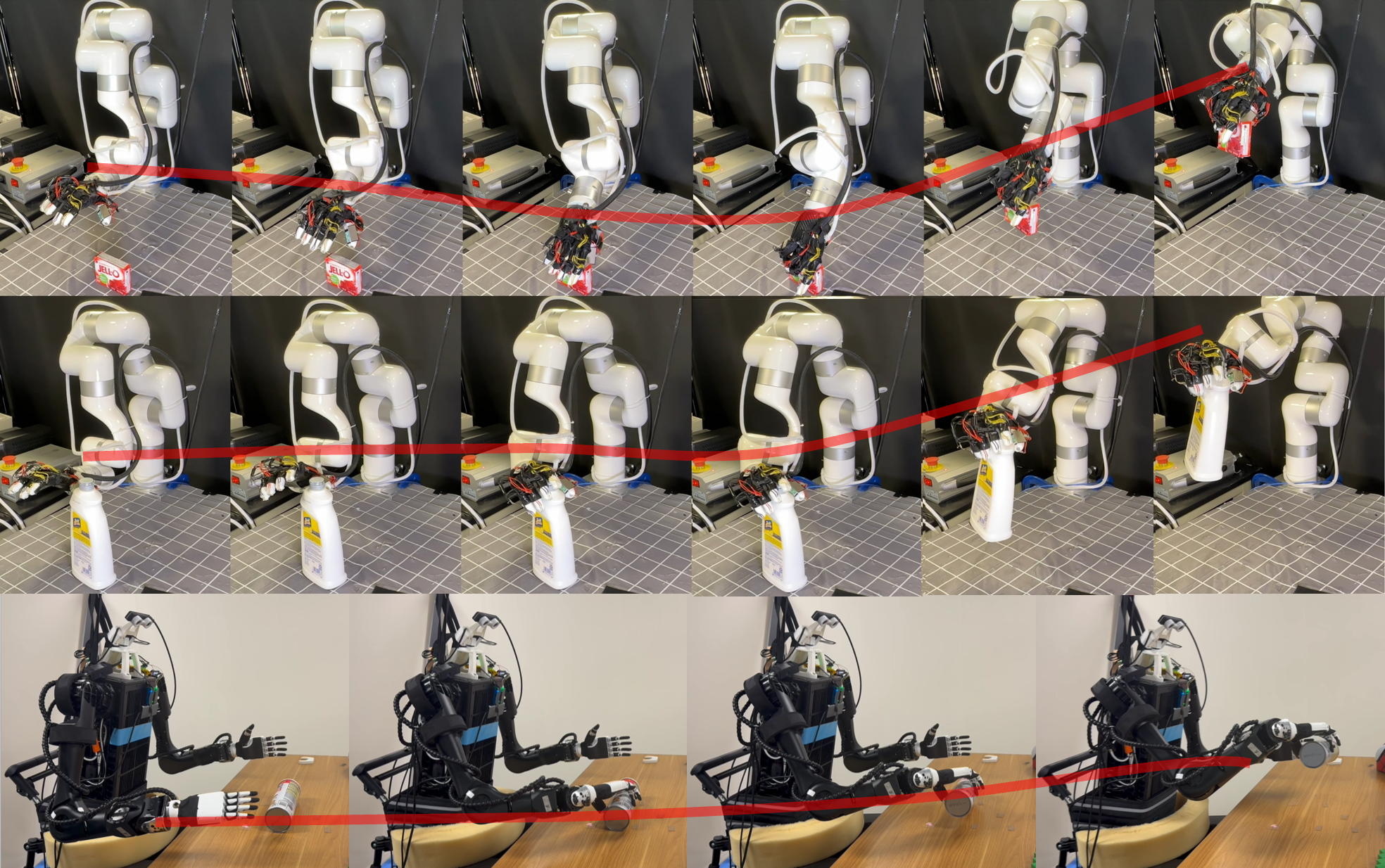} 
  \caption{We directly deploy the predicted pose to demonstrate the effectiveness of the proposed method in the real world.}
  \label{fig:real}
\end{figure}

We demonstrate the effectiveness of the proposed method in the real world through direct sim-to-real deployment. We explore two platforms: xArm with an Ability Hand and H1 with an Inspire Hand. We mount a camera in a third-person view for xArm and an egocentric view for H1. The camera pose is calibrated using hand-eye calibration. We then take the partial point cloud observation from the camera as input to the model. Additionally, we sample 128 poses and select a valid inverse kinematics (IK) solution. Finally, we apply motion planning to this pose. The successful grasping trajectories are visualized in Fig.~\ref{fig:real}.

We compare with DexDiffuser~\cite{weng2024DexDiffuser} on \textit{XArm+Ability Hand}.
\begin{wrapfigure}{r}{0.45\linewidth}
    \vspace{-10pt}
    \centering
    \includegraphics[width=1.0\linewidth]{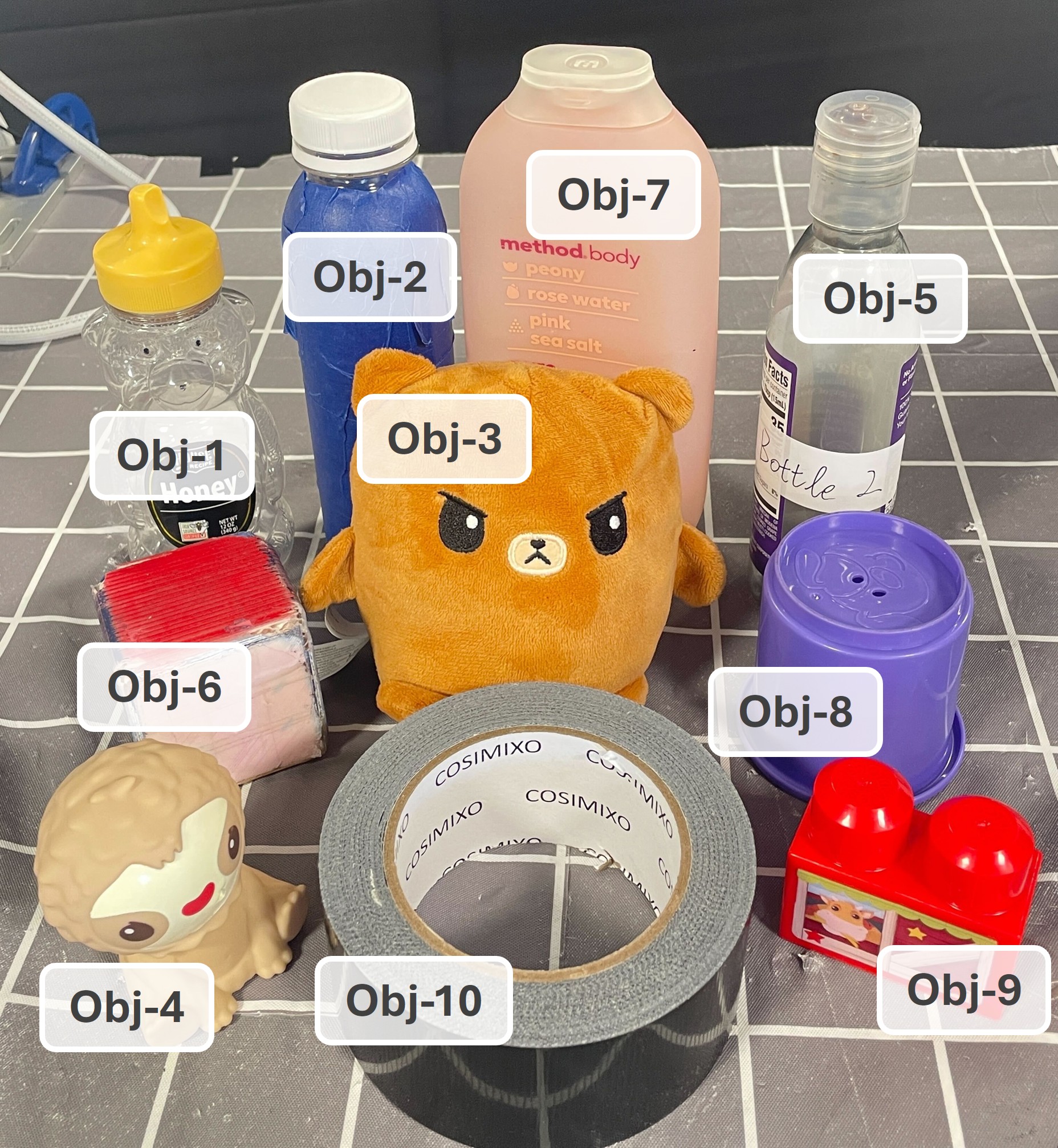}
    \vspace{-15pt}
\end{wrapfigure}
We evaluate 5 trials per object on 10 unseen objects (shown in the figure). Both methods take partial point clouds as input and use motion planning to execute the grasp poses. As DexDiffuser is originally trained for the Allegro Hand, we retrain its model using our dataset and platform. We only retrained the DexSampler module and omitted the DexEvaluator used in their paper. The results below demonstrate the better sim-to-real transfer performance of our proposed method.
\begin{table}[t]
\centering
\small
\setlength{\tabcolsep}{1pt}
\fontsize{8pt}{9pt}\selectfont
\vspace{-8pt}
\resizebox{\linewidth}{!}{%
\begin{tabular}{l cccccccccc c}
\toprule
Method & Obj-1 & Obj-2 & Obj-3 & Obj-4 & Obj-5 & Obj-6 & Obj-7 & Obj-8 & Obj-9 & Obj-10 & Mean \\
\midrule
DexSampler & 2/5 & 3/5 & 5/5 & 3/5 & 1/5 & 4/5 & 2/5 & 5/5 & 1/5 & 3/5 & 58\% \\
Ours & 4/5 & 5/5 & 5/5 & 5/5 & 5/5 & 5/5 & 5/5 & 5/5 & 5/5 & 4/5 & \textbf{96\%} \\
\bottomrule
\end{tabular}%
}
\vspace{-15pt}
\end{table}

\section{Conclusion and Limitations }
\label{sec:conclusion}

In this paper, we present \textbf{\ourdataset}, a synthetic dataset for dexterous hand manipulation, containing 1 billion demonstrations. We introduce an iterative data generation pipeline that integrates optimization techniques with learning-based approaches to efficiently generate manipulation demonstrations. It begins by creating the see dataset using pure optimization methods, which is then used to train our generative model, \textbf{\ourmethod}. The model accelerates the data generation loop by producing a proposal dataset that is further refined through optimization. The refined datasets are subsequently verified and debiased for quality assurance.
In our experiments, we demonstrate that \ourmethod, enhanced with geometric loss, achieves a 22-point improvement over previous state-of-the-art methods on the DexGraspNet benchmark. Additionally, benchmarks for lifting and articulation tasks highlight the effectiveness of both \ourdataset and \ourmethod, showcasing their utility in advancing dexterous hand manipulation research.

\noindent
\textbf{Limitations.} We state several key limitations of our method here: i) Since our method focuses on key-frame action generation, it operates in an open-loop manner when deployed in the real world, making it prone to the sim-to-real gap and control/observation errors. ii) Although the generative model is significantly more efficient than optimization, our method still relies on simulation to filter successful data. The simulation itself remains time-consuming, and reducing its runtime during data generation is a potential direction for future work. iii) Our method mainly considers single-object scene settings. For multi-object scenarios, a stronger vision backbone may be necessary.

\section{Acknowledgment}
This work was supported, in part, by NSF CAREER Award IIS-2240014, Qualcomm Innovation Fellowship, and gifts from Meta.


\bibliographystyle{plainnat}
\bibliography{references}

\begin{thebibliography}{58}
\providecommand{\natexlab}[1]{#1}
\providecommand{\url}[1]{\texttt{#1}}
\expandafter\ifx\csname urlstyle\endcsname\relax
  \providecommand{\doi}[1]{doi: #1}\else
  \providecommand{\doi}{doi: \begingroup \urlstyle{rm}\Url}\fi

\bibitem[Andrychowicz et~al.(2020)Andrychowicz, Baker, Chociej, Jozefowicz, McGrew, Pachocki, Petron, Plappert, Powell, Ray, et~al.]{andrychowicz2020learning}
OpenAI:~Marcin Andrychowicz, Bowen Baker, Maciek Chociej, Rafal Jozefowicz, Bob McGrew, Jakub Pachocki, Arthur Petron, Matthias Plappert, Glenn Powell, Alex Ray, et~al.
\newblock Learning dexterous in-hand manipulation.
\newblock \emph{IJRR}, 2020.

\bibitem[Bao et~al.(2023)Bao, Xu, Qin, and Wang]{bao2023dexart}
Chen Bao, Helin Xu, Yuzhe Qin, and Xiaolong Wang.
\newblock Dexart: Benchmarking generalizable dexterous manipulation with articulated objects.
\newblock In \emph{CVPR}, 2023.

\bibitem[Black et~al.(2024)Black, Brown, Driess, Esmail, Equi, Finn, Fusai, Groom, Hausman, Ichter, et~al.]{black2024pi_0}
Kevin Black, Noah Brown, Danny Driess, Adnan Esmail, Michael Equi, Chelsea Finn, Niccolo Fusai, Lachy Groom, Karol Hausman, Brian Ichter, et~al.
\newblock $\pi 0$: A vision-language-action flow model for general robot control.
\newblock \emph{arXiv}, 2024.

\bibitem[Brahmbhatt et~al.(2019{\natexlab{a}})Brahmbhatt, Ham, Kemp, and Hays]{brahmbhatt2019contactdb}
Samarth Brahmbhatt, Cusuh Ham, Charles~C Kemp, and James Hays.
\newblock Contactdb: Analyzing and predicting grasp contact via thermal imaging.
\newblock In \emph{CVPR}, 2019{\natexlab{a}}.

\bibitem[Brahmbhatt et~al.(2019{\natexlab{b}})Brahmbhatt, Handa, Hays, and Fox]{brahmbhatt2019contactgrasp}
Samarth Brahmbhatt, Ankur Handa, James Hays, and Dieter Fox.
\newblock Contactgrasp: Functional multi-finger grasp synthesis from contact.
\newblock In \emph{IROS}, 2019{\natexlab{b}}.

\bibitem[Brahmbhatt et~al.(2020)Brahmbhatt, Tang, Twigg, Kemp, and Hays]{brahmbhatt2020contactpose}
Samarth Brahmbhatt, Chengcheng Tang, Christopher~D Twigg, Charles~C Kemp, and James Hays.
\newblock Contactpose: A dataset of grasps with object contact and hand pose.
\newblock In \emph{ECCV}, 2020.

\bibitem[Chao et~al.(2021)Chao, Yang, Xiang, Molchanov, Handa, Tremblay, Narang, Van~Wyk, Iqbal, Birchfield, et~al.]{chao2021dexycb}
Yu-Wei Chao, Wei Yang, Yu~Xiang, Pavlo Molchanov, Ankur Handa, Jonathan Tremblay, Yashraj~S Narang, Karl Van~Wyk, Umar Iqbal, Stan Birchfield, et~al.
\newblock Dexycb: A benchmark for capturing hand grasping of objects.
\newblock In \emph{CVPR}, 2021.

\bibitem[Chen et~al.(2024{\natexlab{a}})Chen, Chen, Zhang, and Wang]{chen2023task}
Jiayi Chen, Yuxing Chen, Jialiang Zhang, and He~Wang.
\newblock Task-oriented dexterous grasp synthesis via differentiable grasp wrench boundary estimator.
\newblock \emph{IROS}, 2024{\natexlab{a}}.

\bibitem[Chen et~al.(2024{\natexlab{b}})Chen, Wang, Yang, and Liu]{chen2024object}
Yuanpei Chen, Chen Wang, Yaodong Yang, and Karen Liu.
\newblock Object-centric dexterous manipulation from human motion data.
\newblock In \emph{CoRL}, 2024{\natexlab{b}}.

\bibitem[Chen et~al.(2022)Chen, Van~Wyk, Chao, Yang, Mousavian, Gupta, and Fox]{chen2022dextransfer}
Zoey~Qiuyu Chen, Karl Van~Wyk, Yu-Wei Chao, Wei Yang, Arsalan Mousavian, Abhishek Gupta, and Dieter Fox.
\newblock Dextransfer: Real world multi-fingered dexterous grasping with minimal human demonstrations.
\newblock \emph{RSS IL workshop}, 2022.

\bibitem[Chi et~al.(2024)Chi, Xu, Pan, Cousineau, Burchfiel, Feng, Tedrake, and Song]{chi2024universal}
Cheng Chi, Zhenjia Xu, Chuer Pan, Eric Cousineau, Benjamin Burchfiel, Siyuan Feng, Russ Tedrake, and Shuran Song.
\newblock Universal manipulation interface: In-the-wild robot teaching without in-the-wild robots.
\newblock In \emph{RSS}, 2024.

\bibitem[Christen et~al.(2024)Christen, Feng, Yang, Chao, Hilliges, and Song]{christen2024synh2r}
Sammy Christen, Lan Feng, Wei Yang, Yu-Wei Chao, Otmar Hilliges, and Jie Song.
\newblock Synh2r: Synthesizing hand-object motions for learning human-to-robot handovers.
\newblock In \emph{ICRA}, 2024.

\bibitem[Company(2005)]{shadowhand2005}
Shadow~Robot Company.
\newblock Shadow hand, 2005.
\newblock URL \url{https://www.shadowrobot.com/dexterous-hand-series/}.

\bibitem[Corona et~al.(2020)Corona, Pumarola, Alenya, Moreno-Noguer, and Rogez]{corona2020ganhand}
Enric Corona, Albert Pumarola, Guillem Alenya, Francesc Moreno-Noguer, and Gr{\'e}gory Rogez.
\newblock Ganhand: Predicting human grasp affordances in multi-object scenes.
\newblock In \emph{CVPR}, 2020.

\bibitem[Dai et~al.(2018)Dai, Majumdar, and Tedrake]{dai2018synthesis}
Hongkai Dai, Anirudha Majumdar, and Russ Tedrake.
\newblock Synthesis and optimization of force closure grasps via sequential semidefinite programming.
\newblock \emph{ISRR}, 2018.

\bibitem[Fan et~al.(2023)Fan, Taheri, Tzionas, Kocabas, Kaufmann, Black, and Hilliges]{fan2023arctic}
Zicong Fan, Omid Taheri, Dimitrios Tzionas, Muhammed Kocabas, Manuel Kaufmann, Michael~J. Black, and Otmar Hilliges.
\newblock {ARCTIC}: A dataset for dexterous bimanual hand-object manipulation.
\newblock In \emph{CVPR}, 2023.

\bibitem[Ferrari et~al.(1992)Ferrari, Canny, et~al.]{ferrari1992planning}
Carlo Ferrari, John~F Canny, et~al.
\newblock Planning optimal grasps.
\newblock In \emph{ICRA}, 1992.

\bibitem[Fu et~al.(2024)Fu, Zhao, and Finn]{fu2024mobile}
Zipeng Fu, Tony~Z. Zhao, and Chelsea Finn.
\newblock Mobile aloha: Learning bimanual mobile manipulation with low-cost whole-body teleoperation.
\newblock In \emph{CoRL}, 2024.

\bibitem[Hampali et~al.(2020)Hampali, Rad, Oberweger, and Lepetit]{hampali2020honnotate}
Shreyas Hampali, Mahdi Rad, Markus Oberweger, and Vincent Lepetit.
\newblock Honnotate: A method for 3d annotation of hand and object poses.
\newblock In \emph{CVPR}, 2020.

\bibitem[Hasson et~al.(2019)Hasson, Varol, Tzionas, Kalevatykh, Black, Laptev, and Schmid]{hasson2019learning}
Yana Hasson, Gul Varol, Dimitrios Tzionas, Igor Kalevatykh, Michael~J Black, Ivan Laptev, and Cordelia Schmid.
\newblock Learning joint reconstruction of hands and manipulated objects.
\newblock In \emph{CVPR}, 2019.

\bibitem[Jiang et~al.(2021{\natexlab{a}})Jiang, Liu, Wang, and Wang]{jiang2021graspTTA}
Hanwen Jiang, Shaowei Liu, Jiashun Wang, and Xiaolong Wang.
\newblock Hand-object contact consistency reasoning for human grasps generation.
\newblock In \emph{ICCV}, 2021{\natexlab{a}}.

\bibitem[Jiang et~al.(2021{\natexlab{b}})Jiang, Liu, Wang, and Wang]{jiang2021hand}
Hanwen Jiang, Shaowei Liu, Jiashun Wang, and Xiaolong Wang.
\newblock Hand-object contact consistency reasoning for human grasps generation.
\newblock In \emph{ICCV}, 2021{\natexlab{b}}.

\bibitem[Liu et~al.(2020)Liu, Pan, Xu, Ganguly, and Manocha]{Liu2020DDG}
Min Liu, Zherong Pan, Kai Xu, Kanishka Ganguly, and Dinesh Manocha.
\newblock Deep differentiable grasp planner for high-dof grippers.
\newblock In \emph{RSS}, 2020.

\bibitem[Liu et~al.(2023)Liu, Zhou, Yang, Gupta, and Wang]{liu2023contactgen}
Shaowei Liu, Yang Zhou, Jimei Yang, Saurabh Gupta, and Shenlong Wang.
\newblock Contactgen: Generative contact modeling for grasp generation.
\newblock In \emph{ICCV}, 2023.

\bibitem[Liu et~al.(2021)Liu, Liu, Jiao, Zhu, and Zhu]{liu2021synthesizing}
Tengyu Liu, Zeyu Liu, Ziyuan Jiao, Yixin Zhu, and Song-Chun Zhu.
\newblock Synthesizing diverse and physically stable grasps with arbitrary hand structures using differentiable force closure estimator.
\newblock \emph{RA-L}, 2021.

\bibitem[Liu et~al.(2024)Liu, Yang, Wang, Wu, Wang, Yao, Schwertfeger, Yang, Wang, Yu, He, and Ma]{DBLP:conf/ijcai/LiuYWWWYSYWY0M24}
Yumeng Liu, Yaxun Yang, Youzhuo Wang, Xiaofei Wu, Jiamin Wang, Yichen Yao, S{\"{o}}ren Schwertfeger, Sibei Yang, Wenping Wang, Jingyi Yu, Xuming He, and Yuexin Ma.
\newblock Realdex: Towards human-like grasping for robotic dexterous hand.
\newblock In \emph{IJCAI}, 2024.

\bibitem[Lu et~al.(2024)Lu, Kang, Li, Liu, Yang, Huang, and Hua]{lu2023ugg}
Jiaxin Lu, Hao Kang, Haoxiang Li, Bo~Liu, Yiding Yang, Qixing Huang, and Gang Hua.
\newblock Ugg: Unified generative grasping.
\newblock In \emph{ECCV}, 2024.

\bibitem[Lu et~al.(2020)Lu, Van~der Merwe, and Hermans]{lu2020multi}
Qingkai Lu, Mark Van~der Merwe, and Tucker Hermans.
\newblock Multi-fingered active grasp learning.
\newblock In \emph{IROS}, 2020.

\bibitem[Lum et~al.(2024)Lum, Matak, Makoviychuk, Handa, Allshire, Hermans, Ratliff, and Wyk]{DBLP:conf/corl/LumMMHAHRW24}
Tyler Ga~Wei Lum, Martin Matak, Viktor Makoviychuk, Ankur Handa, Arthur Allshire, Tucker Hermans, Nathan~D. Ratliff, and Karl~Van Wyk.
\newblock Dextrah-g: Pixels-to-action dexterous arm-hand grasping with geometric fabrics.
\newblock In \emph{CoRL}, 2024.

\bibitem[Lundell et~al.(2021)Lundell, Verdoja, and Kyrki]{lundell2021ddgc}
Jens Lundell, Francesco Verdoja, and Ville Kyrki.
\newblock Ddgc: Generative deep dexterous grasping in clutter.
\newblock \emph{RA-L}, 2021.

\bibitem[Macklin(2022)]{warp2022}
Miles Macklin.
\newblock Warp: A high-performance python framework for gpu simulation and graphics.
\newblock \url{https://github.com/nvidia/warp}, March 2022.

\bibitem[Makoviychuk et~al.(2021)Makoviychuk, Wawrzyniak, Guo, Lu, Storey, Macklin, Hoeller, Rudin, Allshire, Handa, et~al.]{makoviychuk2021isaac}
Viktor Makoviychuk, Lukasz Wawrzyniak, Yunrong Guo, Michelle Lu, Kier Storey, Miles Macklin, David Hoeller, Nikita Rudin, Arthur Allshire, Ankur Handa, et~al.
\newblock Isaac gym: High performance gpu-based physics simulation for robot learning.
\newblock In \emph{NeurIPS Datasets and Benchmarks}, 2021.

\bibitem[Nagabandi et~al.(2020)Nagabandi, Konolige, Levine, and Kumar]{nagabandi2020deep}
Anusha Nagabandi, Kurt Konolige, Sergey Levine, and Vikash Kumar.
\newblock Deep dynamics models for learning dexterous manipulation.
\newblock In \emph{CoRL}, 2020.

\bibitem[Ponce et~al.(1993)Ponce, Sullivan, Boissonnat, and Merlet]{ponce1993characterizing}
Jean Ponce, Steve Sullivan, J-D Boissonnat, and J-P Merlet.
\newblock On characterizing and computing three-and four-finger force-closure grasps of polyhedral objects.
\newblock In \emph{ICRA}, 1993.

\bibitem[Ponce et~al.(1997)Ponce, Sullivan, Sudsang, Boissonnat, and Merlet]{ponce1997computing}
Jean Ponce, Steve Sullivan, Attawith Sudsang, Jean-Daniel Boissonnat, and Jean-Pierre Merlet.
\newblock On computing four-finger equilibrium and force-closure grasps of polyhedral objects.
\newblock \emph{IJRR}, 1997.

\bibitem[Prattichizzo et~al.(2012)Prattichizzo, Malvezzi, Gabiccini, and Bicchi]{prattichizzo2012manipulability}
Domenico Prattichizzo, Monica Malvezzi, Marco Gabiccini, and Antonio Bicchi.
\newblock On the manipulability ellipsoids of underactuated robotic hands with compliance.
\newblock \emph{RAS}, 2012.

\bibitem[Qi et~al.(2017)Qi, Su, Mo, and Guibas]{qi2017pointnet}
Charles~R Qi, Hao Su, Kaichun Mo, and Leonidas~J Guibas.
\newblock Pointnet: Deep learning on point sets for 3d classification and segmentation.
\newblock In \emph{CVPR}, 2017.

\bibitem[Qin et~al.(2022)Qin, Wu, Liu, Jiang, Yang, Fu, and Wang]{qin2022dexmv}
Yuzhe Qin, Yueh-Hua Wu, Shaowei Liu, Hanwen Jiang, Ruihan Yang, Yang Fu, and Xiaolong Wang.
\newblock Dexmv: Imitation learning for dexterous manipulation from human videos.
\newblock In \emph{ECCV}, 2022.

\bibitem[Rodriguez et~al.(2012)Rodriguez, Mason, and Ferry]{rodriguez2012caging}
Alberto Rodriguez, Matthew~T Mason, and Steve Ferry.
\newblock From caging to grasping.
\newblock \emph{IJRR}, 2012.

\bibitem[Rosales et~al.(2012)Rosales, Su{\'a}rez, Gabiccini, and Bicchi]{rosales2012synthesis}
Carlos Rosales, Ra{\'u}l Su{\'a}rez, Marco Gabiccini, and Antonio Bicchi.
\newblock On the synthesis of feasible and prehensile robotic grasps.
\newblock In \emph{ICRA}, 2012.

\bibitem[Shao et~al.(2020)Shao, Ferreira, Jorda, Nambiar, Luo, Solowjow, Ojea, Khatib, and Bohg]{shao2020unigrasp}
Lin Shao, Fabio Ferreira, Mikael Jorda, Varun Nambiar, Jianlan Luo, Eugen Solowjow, Juan~Aparicio Ojea, Oussama Khatib, and Jeannette Bohg.
\newblock Unigrasp: Learning a unified model to grasp with multifingered robotic hands.
\newblock \emph{RA-L}, 2020.

\bibitem[Singh et~al.(2024)Singh, Allshire, Handa, Ratliff, and Van~Wyk]{singh2024dextrah}
Ritvik Singh, Arthur Allshire, Ankur Handa, Nathan Ratliff, and Karl Van~Wyk.
\newblock Dextrah-rgb: Visuomotor policies to grasp anything with dexterous hands.
\newblock \emph{arXiv preprint arXiv:2412.01791}, 2024.

\bibitem[Sundaralingam et~al.(2023)Sundaralingam, Hari, Fishman, Garrett, Van~Wyk, Blukis, Millane, Oleynikova, Handa, Ramos, et~al.]{sundaralingam2023curobo}
Balakumar Sundaralingam, Siva Kumar~Sastry Hari, Adam Fishman, Caelan Garrett, Karl Van~Wyk, Valts Blukis, Alexander Millane, Helen Oleynikova, Ankur Handa, Fabio Ramos, et~al.
\newblock Curobo: Parallelized collision-free robot motion generation.
\newblock In \emph{ICRA}, 2023.

\bibitem[Tao et~al.(2024)Tao, Xiang, Shukla, Qin, Hinrichsen, Yuan, Bao, Lin, Liu, kai Chan, Gao, Li, Mu, Xiao, Gurha, Huang, Calandra, Chen, Luo, and Su]{taomaniskill3}
Stone Tao, Fanbo Xiang, Arth Shukla, Yuzhe Qin, Xander Hinrichsen, Xiaodi Yuan, Chen Bao, Xinsong Lin, Yulin Liu, Tse kai Chan, Yuan Gao, Xuanlin Li, Tongzhou Mu, Nan Xiao, Arnav Gurha, Zhiao Huang, Roberto Calandra, Rui Chen, Shan Luo, and Hao Su.
\newblock Maniskill3: Gpu parallelized robotics simulation and rendering for generalizable embodied ai.
\newblock \emph{arXiv}, 2024.

\bibitem[Turpin et~al.(2022)Turpin, Wang, Heiden, Chen, Macklin, Tsogkas, Dickinson, and Garg]{turpin2022grasp}
Dylan Turpin, Liquan Wang, Eric Heiden, Yun-Chun Chen, Miles Macklin, Stavros Tsogkas, Sven Dickinson, and Animesh Garg.
\newblock Grasp’d: Differentiable contact-rich grasp synthesis for multi-fingered hands.
\newblock In \emph{ECCV}, 2022.

\bibitem[Turpin et~al.(2023)Turpin, Zhong, Zhang, Zhu, Heiden, Macklin, Tsogkas, Dickinson, and Garg]{turpin2023fast}
Dylan Turpin, Tao Zhong, Shutong Zhang, Guanglei Zhu, Eric Heiden, Miles Macklin, Stavros Tsogkas, Sven Dickinson, and Animesh Garg.
\newblock Fast-grasp'd: Dexterous multi-finger grasp generation through differentiable simulation.
\newblock In \emph{ICRA}, 2023.

\bibitem[Wang et~al.(2023)Wang, Zhang, Chen, Xu, Li, Liu, and Wang]{wang2023dexgraspnet}
Ruicheng Wang, Jialiang Zhang, Jiayi Chen, Yinzhen Xu, Puhao Li, Tengyu Liu, and He~Wang.
\newblock Dexgraspnet: A large-scale robotic dexterous grasp dataset for general objects based on simulation.
\newblock In \emph{ICRA}, 2023.

\bibitem[Weng et~al.(2024)Weng, Lu, Kragic, and Lundell]{weng2024DexDiffuser}
Zehang Weng, Haofei Lu, Danica Kragic, and Jens Lundell.
\newblock Dexdiffuser: Generating dexterous grasps with diffusion models.
\newblock \emph{RA-L}, 2024.

\bibitem[Wu et~al.(2022)Wu, Guo, and Liu]{wu2022learning}
Albert Wu, Michelle Guo, and C~Karen Liu.
\newblock Learning diverse and physically feasible dexterous grasps with generative model and bilevel optimization.
\newblock \emph{CoRL}, 2022.

\bibitem[Xiang et~al.(2020)Xiang, Qin, Mo, Xia, Zhu, Liu, Liu, Jiang, Yuan, Wang, et~al.]{xiang2020sapien}
Fanbo Xiang, Yuzhe Qin, Kaichun Mo, Yikuan Xia, Hao Zhu, Fangchen Liu, Minghua Liu, Hanxiao Jiang, Yifu Yuan, He~Wang, et~al.
\newblock Sapien: A simulated part-based interactive environment.
\newblock In \emph{CVPR}, 2020.

\bibitem[Xu et~al.(2023)Xu, Wan, Zhang, Liu, Shan, Shen, Wang, Geng, Weng, Chen, et~al.]{xu2023unidexgrasp}
Yinzhen Xu, Weikang Wan, Jialiang Zhang, Haoran Liu, Zikang Shan, Hao Shen, Ruicheng Wang, Haoran Geng, Yijia Weng, Jiayi Chen, et~al.
\newblock Unidexgrasp: Universal robotic dexterous grasping via learning diverse proposal generation and goal-conditioned policy.
\newblock In \emph{CVPR}, 2023.

\bibitem[Yang et~al.(2021)Yang, Zhan, Li, Xu, Li, and Lu]{yang2021cpf}
Lixin Yang, Xinyu Zhan, Kailin Li, Wenqiang Xu, Jiefeng Li, and Cewu Lu.
\newblock Cpf: Learning a contact potential field to model the hand-object interaction.
\newblock In \emph{ICCV}, 2021.

\bibitem[Ye et~al.(2023)Ye, Wang, Huang, Qin, and Wang]{ye2023learning}
Jianglong Ye, Jiashun Wang, Binghao Huang, Yuzhe Qin, and Xiaolong Wang.
\newblock Learning continuous grasping function with a dexterous hand from human demonstrations.
\newblock \emph{RA-L}, 2023.

\bibitem[Zhang et~al.(2024)Zhang, Liu, Li, Yu, Geng, Ding, Chen, and Wang]{zhang2024dexgraspnet}
Jialiang Zhang, Haoran Liu, Danshi Li, XinQiang Yu, Haoran Geng, Yufei Ding, Jiayi Chen, and He~Wang.
\newblock Dexgraspnet 2.0: Learning generative dexterous grasping in large-scale synthetic cluttered scenes.
\newblock In \emph{CoRL}, 2024.

\bibitem[Zhao et~al.(2023)Zhao, Kumar, Levine, and Finn]{DBLP:conf/rss/ZhaoKLF23}
Tony~Z. Zhao, Vikash Kumar, Sergey Levine, and Chelsea Finn.
\newblock Learning fine-grained bimanual manipulation with low-cost hardware.
\newblock In Kostas~E. Bekris, Kris Hauser, Sylvia~L. Herbert, and Jingjin Yu, editors, \emph{RSS}, 2023.

\bibitem[Zhao et~al.(2024{\natexlab{a}})Zhao, Tompson, Driess, Florence, Ghasemipour, Finn, and Wahid]{zhao2024aloha}
Tony~Z Zhao, Jonathan Tompson, Danny Driess, Pete Florence, Kamyar Ghasemipour, Chelsea Finn, and Ayzaan Wahid.
\newblock Aloha unleashed: A simple recipe for robot dexterity.
\newblock In \emph{CoRL}, 2024{\natexlab{a}}.

\bibitem[Zhao et~al.(2024{\natexlab{b}})Zhao, Chen, Schneider, Gao, Kannala, Sch{\"o}lkopf, Pajarinen, and B{\"u}chler]{zhao2024rp1m}
Yi~Zhao, Le~Chen, Jan Schneider, Quankai Gao, Juho Kannala, Bernhard Sch{\"o}lkopf, Joni Pajarinen, and Dieter B{\"u}chler.
\newblock Rp1m: A large-scale motion dataset for piano playing with bi-manual dexterous robot hands.
\newblock \emph{arXiv preprint arXiv:2408.11048}, 2024{\natexlab{b}}.

\bibitem[Zheng and Chew(2009)]{zheng2009distance}
Yu~Zheng and Chee-Meng Chew.
\newblock Distance between a point and a convex cone in $ n $-dimensional space: Computation and applications.
\newblock \emph{T-RO}, 2009.

\end{thebibliography}

\clearpage
\def\maketitlesupplementary
   {
   \newpage
       \twocolumn[
        \centering
        \Large
        \textbf{\ourdataset: Learning with 1B Demonstrations for Dexterous Manipulation}\\
        \vspace{0.5em} Appendix \\
        \vspace{1.0em}
       ] 
   }
\maketitlesupplementary

\setcounter{section}{0}

\section{Grasping Synthesis Evaluation Details}
We evaluate the performance of the proposed generative model, \ourmethod, using the DexGraspNet~\cite{wang2023dexgraspnet} benchmark. This benchmark includes 5355 objects, with each object associated with approximately 200 grasps across five scales: $\{0.06, 0.08, 0.1, 0.12, 0.15\}$. We follow the train/test splits provided by the benchmark, which divide it into 4229 objects for training and 1126 objects for testing. We train \ourmethod\ only using the training data provided by the benchmark. The results of DDG~\cite{Liu2020DDG}, GraspTTA~\cite{jiang2021graspTTA}, UDG~\cite{xu2023unidexgrasp} and UGG~\cite{lu2023ugg} are taken from the UGG~\cite{lu2023ugg} paper.

In the evaluation, we adhere to the metrics established in DexGraspNet~\cite{wang2023dexgraspnet}, which are divided into two categories: Quality (Success Rate, $Q_1$-score, Penetration) and Diversity (H mean and H std). We detail each metric here.

\noindent
\textbf{Quality Metrics}. i). \textit{Success rate ($\%$)}: A grasp is considered successful if it withstands at least one of the six gravity directions in the Isaac Gym simulator~\cite{makoviychuk2021isaac} and maintains a maximal penetration depth of less than 0.5 cm. ii). \textit{$Q_1$-score}: $Q_1$~\cite{ferrari1992planning} is the radius of the inscribed sphere of the ConvexHull ($\cup_i w_i$), indicating the norm of the smallest wrench that can destabilize the grasp. The contact threshold is set to 1 cm. iii). \textit{Maximal Penetration Depth (cm)}: This metric measures the penetration depth from the object point cloud to the hand meshes.

\noindent
\textbf{Diversity Metrics}. \textit{Diversity} in the DexGraspNet benchmark is evaluated using joint angle entropy. The range of joint motion is divided into $10000$ bins\footnote{The value $10000$ is taken from the UGG implementation, and the entropy results obtained using this value are consistent with those reported in \cite{wang2023dexgraspnet,lu2023ugg}. However, the descriptions provided in \cite{wang2023dexgraspnet,lu2023ugg} are inaccurate.}, and all generated samples are used to estimate a probability distribution. The entropy is calculated based on this distribution. The reported values in main text refer to the mean and standard deviation across all joints (H mean and H std).

\section{Benchmark Details}
The proposed \ourdataset\ dataset contains 3,491 objects for training and 933 objects for testing in the lifting task, as well as 43 articulation objects for training and 21 articulation objects for testing in the articulation task.

The lifitng and articulation tasks are conducted in the ManiSkill~\cite{taomaniskill3}/SAPIEN~\cite{xiang2020sapien} simulation environments. The physical parameters are listed in Tab.~\ref{tab:phy-params}.

\begin{table}[h]
\small
\centering
\setlength{\tabcolsep}{5pt}
\fontsize{9pt}{10pt}\selectfont
\begin{tabular}{l c}
\toprule
Parameter & Value \\
\midrule
Object Mass & 0.1 kg \\
Simulation Frequency & 100 \\
Control Frequency & 25 \\
Contact Offset & 0.001 \\
Solver Position Iterations & 30 \\
\bottomrule
\end{tabular}
\caption{Physical parameters for the simulation.}
\label{tab:phy-params} 
\end{table}

\begin{figure}[h]
\centering
  \includegraphics[width=.9\linewidth]{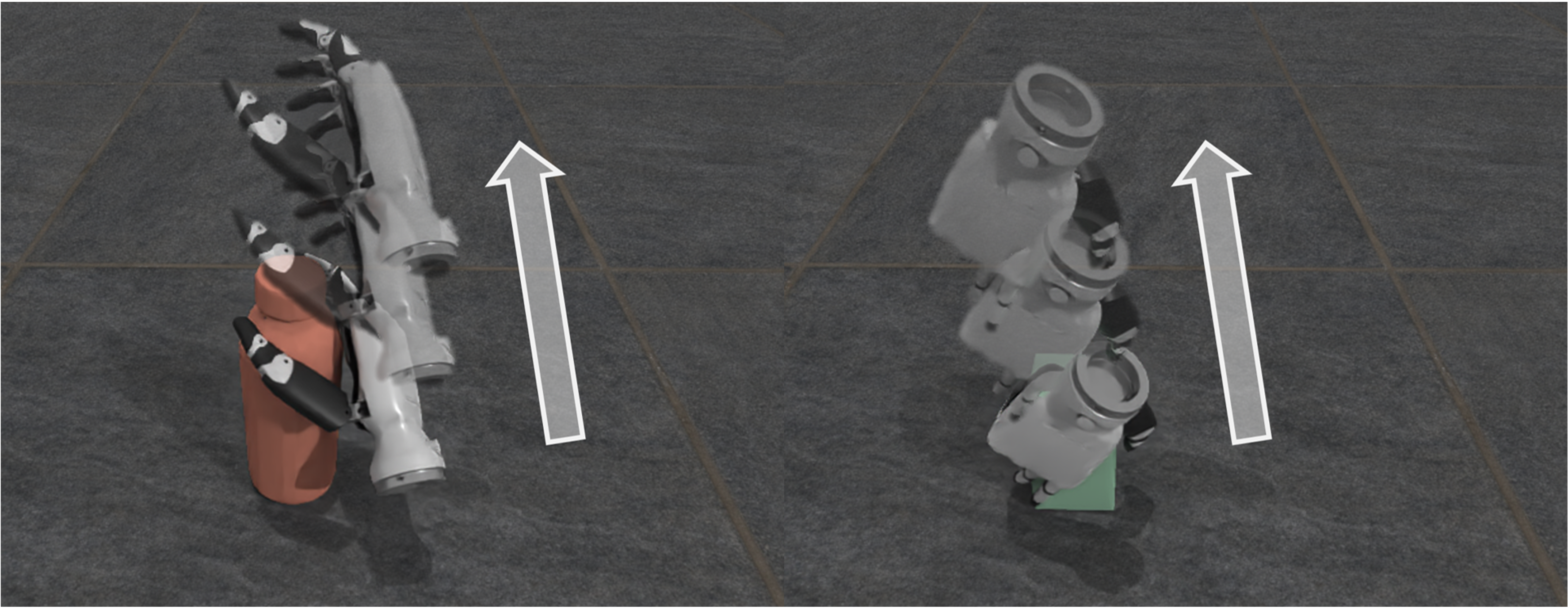} 
  \vspace{-.6em}
  \caption{\textbf{Lifting trajectory} from \ourdataset\ dataset.}
  \vspace{-.6em}
  \label{fig:lifting-task}
\end{figure}

\noindent
\textbf{Lifting Task Definition}. The hand begins from a randomly sampled pose and joint configuration. The goal of the lifting task is to reach an object placed on a plane (table), grasp it, and lift it to a height of 0.4 m. Additionally, the task requires at least two fingers to maintain contact with the object in the final frame. We show example trajectories from \ourdataset\ for the lifting task in Fig.~\ref{fig:lifting-task}.

\begin{figure}[h]
\centering
  \includegraphics[width=0.9\linewidth]{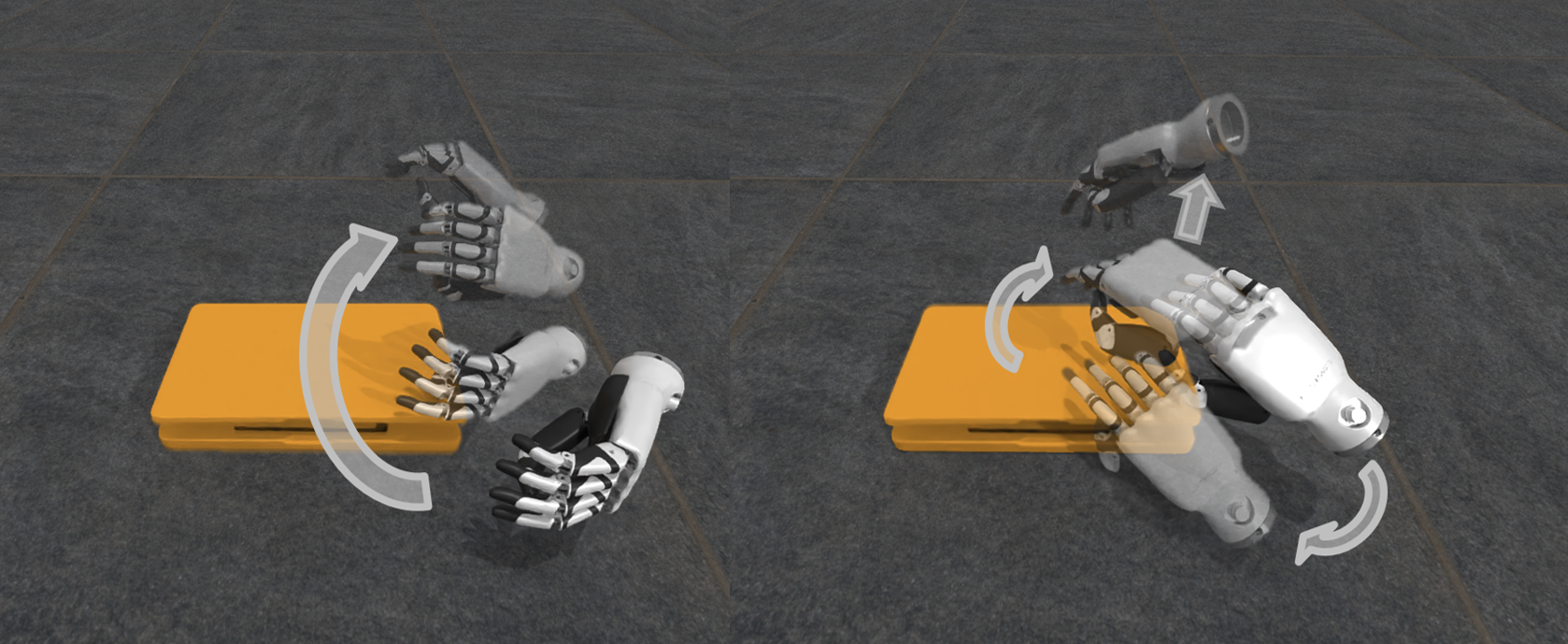}
  \vspace{-2pt}
  \caption{\textbf{Articualtion trajectory} from \ourdataset\ dataset.}
  \vspace{-0.5em}
  \label{fig:arti-task}
\end{figure}

\noindent
\textbf{Articulation Task Definition}.
The hand begins from a randomly sampled pose and joint configuration.
The goal of the articulation task is to reach the interactable link of an articulated structure (e.g., the top link of a laptop) and open it to increase its joint angle by $0.5$ radians. Additionally, the task requires at least two fingers to maintain contact with the object in the final frame.
We show example trajectories from \ourdataset\ for the lifting task in Fig.~\ref{fig:arti-task}.

\section{Iterative Data Engine Details}

The Iterative Data Engine begins with a controllable optimization algorithm for hand pose and trajectory optimization. Both the grasping synthesis and motion planning can be formulated as optimization problems~\cite{liu2021synthesizing,wang2023dexgraspnet,sundaralingam2023curobo}. We first describe the pure optimization-based data generation process for both grasping and motion planning.

\subsection{Grasping Synthesis as Optimizaiton}

\noindent
\textbf{Differentiable Force Closure}. Following ~\cite{liu2021synthesizing,wang2023dexgraspnet}, we adopt differentiable force closure estimator as an energy term for grasping optimization. This term encourages a set of contact points to form force closure, can be expressed as
\begin{equation}
    E_{\rm fc}=\|Gc\|_2
\end{equation}
where
$$
G=\begin{bmatrix}
I_3&\cdots&I_3\\
[x_1]_\times&\cdots&[x_n]_\times
\end{bmatrix}
$$
$$
{[x_i]}_\times=\begin{bmatrix}0&-x_i^{(z)}&x_i^{(y)}\\x_i^{(z)}&0&-x_i^{(x)}\\-x_i^{(y)}&x_i^{(x)}&0\end{bmatrix}
$$

Here $x$ represents $n = 4$ contact points, $c\in \mathbb{R}^{n\times 3}$ represents  the contact normal vectors, which can be computed given object mesh $O$ and $x$. As in DexGraspNet, the contact points are randomly chosen from a predefined set of candidate contact points at each iteration.

\noindent
\textbf{SDF Energy}. To prevent penetration, ~\cite{wang2023dexgraspnet,jiang2021graspTTA} represent objects as point clouds and compute the penetration distance between object point clouds and hand link meshes. In contrast, we represent objects as meshes and hand links as spheres, calculate the sphere-mesh signed distance function (SDF), and leverage a BVH structure to accelerate computations. 

\begin{equation}
    E_{\texttt{sdf}} = \max\left(-\min_{s} \left( \text{SDF}(s_{\text{center}}) + s_{\text{radius}} \right), 0\right)
\end{equation}

Here, $s$ denotes the sphere, $s_{\text{center}} \in \mathbb{R}^{n\times 3}$ is the 3D position of the sphere center, and $s_{\text{radius}} \in \mathbb{R}^+$ is the sphere radius. $\text{SDF}(s_{\text{center}})$ is the SDF query operation (inside is negative). We implement this term using warp-lang~\cite{warp2022}. The resulting SDF term is far more accurate and efficient.

\noindent
\textbf{Other Energy Terms.} We follow DexGraspNet~\cite{wang2023dexgraspnet} to add contact distance (\(E_{\texttt{D}}\)), self-penetration energy (\(E_{\texttt{S}}\)), and joint limit energy (\(E_{\texttt{J}}\)):

\begin{equation}
\begin{aligned}
    E_{\texttt D} &= \sum_{i} d(x_i, O), \\
    E_{\texttt S} &= \sum \max_{p \ne q}(\delta - d(p, q), 0), \\
    E_{\texttt J} &= \sum_{i} \left(\max(\theta_i - \theta_i^{\max}, 0) + \max(\theta_i^{\min} - \theta_i, 0)\right)
\end{aligned}
\end{equation}

Here, contact distance (\(E_{\texttt{D}}\)) is the distance between predefined contact points $x_i$ and object mesh $O$. Self-penetration energy (\(E_{\texttt{S}}\)) is the threshold  $\delta = 0.02$ minus the distance betwee joints $p, q$. Joint limit energy (\(E_{\texttt{J}}\)) is the deviation for joint angles $\theta$ from their joint limits.

The complete energy function for grasping synthesis is as follows:
\begin{equation}
    E = E_{\texttt{fc}} +  w_{\texttt{sdf}}E_{\rm{sdf}} +  w_{\texttt{D}}E_{\texttt{D}} +  w_{\texttt{S}}E_{\texttt{S}} + w_{\texttt{J}}E_{\texttt{J}},
\end{equation}

The weights are set as follows: $ w_{\texttt{sdf}} = 100.0$, $w_{\texttt{D}} = 100.0$, $w_{\texttt{S}} = 10.0$, $w_{\texttt{J}} = 1.0$.

\subsection{Motion Planning as Optimizaiton}

We formulate motion planning as an optimization problem incorporating SDF energy ($E_{\texttt{sdf}}$), self-penetration energy (\(E_{\texttt{S}}\)), joint limit energy (\(E_{\texttt{J}}\)), and smoothness energy ($E_{\texttt{smooth}}$).  The first three energy terms are extensions of single-frame grasping formulations to multi-frame settings. The last smoothness energy ensures velocity smoothness:

\begin{equation}
E_{\text{smooth}} = {\textstyle \sum_{t=2}^T} \left\| (g_t - g_{t-1}) / \Delta t \right\|^2
\end{equation}

Here, \(g = (T, R, \theta)\) represents the grasp tuple with \(T \in \mathbb{R}^3\) for global translation, \(R \in SO(3)\) for global rotation, and \(\theta \in \mathbb{R}^d\) for joint angles.

\noindent
\textbf{Continuous Euler Angles Optimization}. Free hand is typically implemented using 6 root joints (3 prismatic joints for translation, 3 revolute joints for rotation) in the simulation. The 3 revolute joints are equivalent to intrinsic Euler angles (x-y-z). However, converting from other rotation representations (such as 6D rotation) to intrinsic Euler angles is not necessarily continuous across time steps. To address this, we use an optimization-based approach for rotation conversion. Specifically, we use a smoothness energy term ($E_{\text{smooth}}$) and a rotation difference term, which indicates the rotation angle difference between current rotation and target rotation, to optimize the revolute joint angles over a trajectory.

\subsection{Task-specific Trajectory Optimization}

Both the lifting and articulation demonstrations can be divided into three stages: pre-grasping, grasping, and post-grasping. The grasping stage involves grasp synthesis, while the pre-grasping stage focuses on motion planning.  In the post-grasping stage, lifting requires raising the hand's height (z-value), whereas articulation requires rotating the hand along the axis of the articulated structure. For all demonstrations, we follow this data generation process: i) Grasp synthesis; ii) Sampling the starting pose and perform motion planning; iii) Generating the task-specific post-grasping trajectory.

\section{\ourmethod\ Model Details}

While generative models~\cite{jiang2021graspTTA,liu2023contactgen,lu2023ugg} have been extensively studied for dexterous manipulation, we introduce several differentiable loss terms inspired by optimization-based methods, SDF loss, distance loss, and smoothness loss, and demonstrate that these losses significantly enhance performance. For the SDF loss, due to the lack of cache support for BVH structures in warp-lang, we use point-sphere SDF queries instead of mesh-sphere SDF queries during large-scale training.

The proposed \ourmethod\ is a CVAE architecture with a PointNet encoder. The architecture details and training hyperparameters for \ourmethod\ are provided in Tab.~\ref{tab:model-params}.

\begin{table}[h]
\small
\centering
\setlength{\tabcolsep}{5pt}
\fontsize{9pt}{10pt}\selectfont
\begin{tabular}{l c}
\toprule
Parameter & Value \\
\midrule
Num. Points & 1024 \\
PointNet Layer Sizes & (3, 64, 128, 1024, 256) \\
CVAE In/Out Dim. & $N_{\texttt{frames}} \times N_{\texttt{DOF}}$ \\
CVAE Layer Sizes & (256, 512, 256) \\
Learning rate & 1e-5 \\
MSE Loss Weight & 1.0 \\
KL Loss Weight & 1e-4 \\
SDF Loss Weight & 1e-4 \\
Distance Loss Weight & 1e-4 \\
Smoothness Loss Weight & 1e-5 \\
Optimizer & Adam \\
\bottomrule
\end{tabular}
\caption{Parameters for \ourmethod.}
\vspace{-1em}
\label{tab:model-params} 
\end{table}

\end{document}